\pdfoutput=1
\PassOptionsToPackage{svgnames}{xcolor}

\documentclass[11pt]{article}

\usepackage[final]{acl}

\usepackage{times}
\usepackage{latexsym}
\usepackage{placeins}
\usepackage{enumitem}

\usepackage[T1]{fontenc}

\usepackage[utf8]{inputenc}

\usepackage{microtype}

\usepackage{inconsolata}

\usepackage{graphicx}

\usepackage{url}
\usepackage{graphicx}
\usepackage{subcaption}
\usepackage{caption}
\usepackage{hyperref}
\usepackage{wrapfig} 
\usepackage{float}
\usepackage{titlesec}

\newcommand{\lm}{\ensuremath{M}}
\newcommand{\dataset}{\ensuremath{\mathcal{D}}}

\newcommand{\novel}{\textcolor{orange}{immediate}\xspace}

\newcommand{\retained}{\textcolor{DarkGreen}{retained}\xspace}

\newcommand{\forgotten}{\textcolor{red}{forgotten}\xspace}

\newcommand{\forgetting}{\textcolor{red}{forgetting}\xspace}

\newcommand{\assisted}{\textcolor{blue}{assisted}\xspace}
\newcommand{\Assisted}{\textcolor{blue}{Assisted}\xspace}

\newif\ifshowcomments
\showcommentstrue

\ifshowcomments

\else
\newcommand\todo[1]{}
\newcommand\red[1]{}
\newcommand\cc[1]{}
\newcommand\kl[1]{}
{}

\fi

\usepackage{xspace}

\newcommand{\ngram}{\ensuremath{n}-gram\@\xspace}
\newcommand{\ngrams}{\ensuremath{n}-grams\@\xspace}

\usepackage[textsize=tiny]{todonotes}

\title{Privacy Ripple Effects from Adding or Removing \\ Personal Information in Language Model Training}

\author{
  \textbf{Jaydeep Borkar\textsuperscript{1}},
  \textbf{Matthew Jagielski\textsuperscript{2}},
  \textbf{Katherine Lee\textsuperscript{2}},
  \textbf{Niloofar Mireshghallah\textsuperscript{3}}, \\
  \textbf{David A. Smith\thanks{Equal senior authorship.}\textsuperscript{1}},
  \textbf{Christopher A. Choquette-Choo\footnotemark[1]\textsuperscript{2}} \\
\\
  \textsuperscript{1}Northeastern University,
  \textsuperscript{2}Google DeepMind,
  \textsuperscript{3}University of Washington
\\
 \small{
   \textbf{Correspondence:} \href{mailto:borkar.j@northeastern.edu}{borkar.j@northeastern.edu} and \href{mailto:cchoquette@google.com}{cchoquette@google.com}
 }
}

\begin{document}
\maketitle
\begin{abstract}

Due to the sensitive nature of personally identifiable information (PII), its owners may have the authority to control its inclusion or request its removal from large-language model (LLM) training. 
Beyond this, PII may be added or removed from training datasets due to evolving dataset curation techniques, because they were newly scraped for retraining, or because they were included in a new downstream fine-tuning stage. We find that the amount and ease of PII memorization is a dynamic property of a model that evolves throughout training pipelines and depends on commonly altered design choices. We characterize three such novel phenomena: (1) similar-appearing PII seen later in training can elicit memorization of earlier-seen sequences in what we call \emph{assisted memorization}, and this is a significant factor (in our settings, up to 1/3); (2) adding PII can increase memorization of other PII significantly (in our settings, as much as $\approx\!7.5\times$); and (3) removing PII can lead to other PII being memorized.
Model creators should consider these first- and second-order privacy risks when training models to avoid the risk of new PII regurgitation. \looseness=-1

\end{abstract}

\section{Introduction}

One of the most common methods to adapt large language models like ChatGPT~\citep{achiam2023gpt} and Gemini~\citep{team2023gemini} for specific applications is to fine-tune them on domain-specific datasets.\footnote{See \url{https://platform.openai.com/docs/guides/fine-tuning/when-to-use-fine-tuning} or \url{https://ai.google.dev/gemini-api/docs/model-tuning}} When these datasets contain private or personal data, models may be at risk of memorizing\footnote{We adopt the definition of “memorization” as used at \url{www.genlaw.org/glossary.html}} and regurgitating~\citep{quantifying} this information.
Though it is common to filter out sensitive information\footnote{We focus on PII as a more concrete privacy risk, though note that our results likely also extend to broader types of sensitive information. We thus use these terms interchangeably.} such as PII~\citep{team2024gemma2}, some sensitive information may still remain~\citep{vakili-etal-2022-downstream}.
Moreover, some downstream tasks, such as healthcare, may require PII, making eliminating PII completely from model training datasets challenging.

\begin{figure}[t]
  \includegraphics[width=1.0\columnwidth]{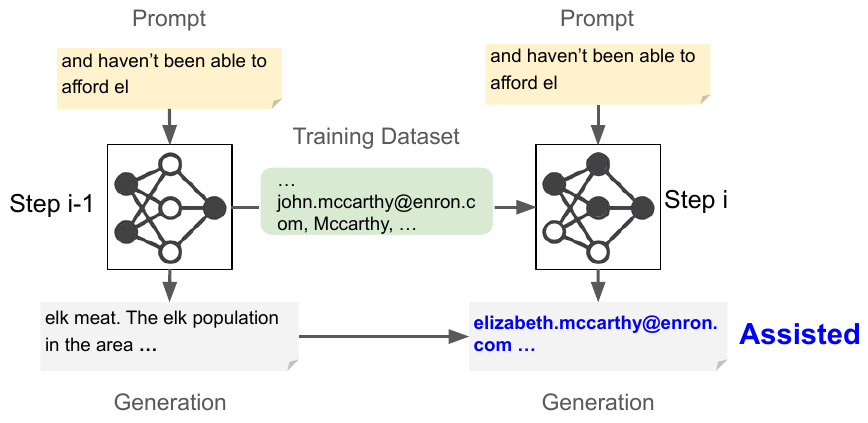}
  
  \caption{
  We explore a phenomenon we call \emph{assisted memorization}, where unique PII that appeared earlier in the training at step $i-1$ and was not extracted at that step becomes extractable at a later step $i$, after fine-tuning on \emph{other} PII.}
 
  \label{fig:taxonomy_assist}
\end{figure}

Modern-day language models deployed in real-world settings are also increasingly dynamic: it is common practice to continually update or retrain them with new and/or additional data~\citep{razdaibiedina2023progressive, ke2023continual, jang2022towards, jin-etal-2022-lifelong-pretraining}, e.g., if new users opt to share their data. 
There may also be data removal requests from existing users under the \emph{right to be forgotten}~\citep{234843}. Here, machine unlearning~\citep{cao2015towards,bourtoule2021machine} is often the proposed solution by enabling post-hoc removal of data (e.g., PII) from neural models after training. 

LLMs are known to memorize and regurgitate personal information and PII~\citep{carlini-extraction,nasr2023scalable}, which is a concrete privacy harm we study. In this literature, little focus has been given to how this may arise dynamically as a part of a machine learning system.
In this work, we study how various actions (continually training on more data, re-training with new data, or re-training after removing data) may influence PII memorization and extraction. We systematically study these operations to determine which improve or worsen the memorization of PII. In particular, we have four \textbf{main contributions}\footnote{Code available at \url{https://github.com/jaydeepborkar/Assisted-Memorization}}:
\begin{enumerate}[itemsep=0.5mm]
    \item We observe the phenomenon of \emph{assisted memorization}: PII may not be memorized immediately after it is seen, but may be memorized later in training (\S\ref{section:assisted_memorization} and Figure~\ref{fig:taxonomy_assist}). We find this is largely influenced by \ngram statistics. 

    \item We propose a taxonomy of types of PII memorization that arise while training an LLM and show how they manifest (\S~\ref{section:continuous_training} and Figure~\ref{fig:taxonomy}). 
    
    \item We observe that introducing new PII into training data may worsen extraction of PII (\S\ref{section:opt-in}).
    
    \item We observe that reducing the PII memorization risks for one individual can worsen these risks for another individual (\S\ref{section:onion}).

   \end{enumerate}

\section{Related Work}

\paragraph{Membership Inference}
is one of the most common privacy attacks on neural models~\citep{shokri2017membership}. Though successful on computer vision models~\citep{yeom2018privacy,salem2018ml,sablayrolles2019white,choquette2021label,carlini2022membership,jagielski2024students}, these attacks are not often successful on LLMs~\citep{duan2024membership} which we study. Thus, and because verbatim extraction poses a stronger privacy risk, we focus on \emph{memorization and extraction}.

\paragraph{Memorization \& Extraction}
studies when a text is trained on and generated by a model. This is widely studied~\citep{secretsharer,carlini-extraction, quantifying, lee-etal-2022-deduplicating, counterfactual, ippolito-etal-2023-preventing,biderman2023emergent, pythia, kudugunta2024madlad,nasr2023scalable, borkar2023learndataleakageunlearning, chang-etal-2023-speak, ozdayi-etal-2023-controlling, compression, decop, wang-etal-2024-unlocking}. These works are often focused on the broad phenomenon, and not the nature of the data, e.g., if it were sensitive as in our work. Relatively fewer works have considered this setting. \citet{huang-etal-2022-large} study if information about specific entites can be extracted; \citet{panda2024teach} study if LLM's can be poisoned to memorize specific PII; \citet{PII-leakage} formalize PII extraction, proposing several attacks and studying the efficacy of various existing defenses; ~\citet{mireshghallah-etal-2022-empirical} and ~\citet{zeng-etal-2024-exploring} study memorization during fine-tuning; and \citet{lehman-etal-2021-bert} found that extracting sensitive data, using simple techniques, from BERT trained on clinical notes was largely unsuccessful. This line of work has become important for practical privacy and memorization audits~\citep{anil2023palm,team2023gemini,dubey-2024-evaluating}, which also often include PII memorization evaluations~\citep{team2023gemini,team2024gemini,team2024gemma,team2024gemma2,team2024codegemma}. 

\paragraph{Dynamics of Memorization.} 
Most related to our work are those exploring memorization throughout training. It is known that language models memorize more as training progresses~\citep{tirumala2022memorization, recite,huang-etal-2024-demystifying} and exhibit forgetting of memorized examples~\citep{jagielski2022measuring}. \citet{biderman2023emergent} found that there is not high correlation between memorized sequences within checkpoints of a training run. \citet{duan2024uncoveringlatentmemoriesassessing} show a similar notion of ``latent memorization'' but that instead  uses Gaussian noise to uncover these latent memories; instead, our ``assisted memorization'' shows this can happen in normal training runs through only naturally occurring text sequences. The literature so far lacks a clear understanding of the complete memorization landscape throughout training. In our work, we provide a complete taxonomy and uncover novel forms of memorization within training dynamics. 

\paragraph{Unlearning} 
Machine unlearning methods have been proposed as an efficient way to erase data from neural networks~\citep{sisa, approx_deletion, unlearning_auditing}. These methods are motivated by scenarios where users may request for their data to be removed from a trained model (possibly due to legislative considerations like GDPR~\citep{Fabbrini_Celeste_2020}). While many techniques have been proposed for machine unlearning, we focus on the simple strategy of retraining without relevant data points which is the current gold standard, though it may not be applicable to all practical scenarios~\citep{cooper2024machineunlearningdoesntthink}. Most related to our work are works that show unlearning can cause additional privacy risks:~\citet{jeopardize} show this can lead to stronger membership inference attacks and~\citet{onioneffect, hayes2024inexact} show that unlearning can increase membership inference accuracy on other training samples.

\section{Experimental Setup}
\label{sec:prelim}
Our goal is to study how memorization of PII manifests during training.\footnote{We do not state or imply [here] that a model ``contains'' its training data in the sense that there is a copy of that data in the model.  Rather, a model memorizes attributes of its training data such that in certain cases it is statistically able to generate such training data when following rules and using information about features of its training data that it does contain.} This includes continual training or fine-tuning setups in \S\ref{section:continuous_training} and re-training or unlearning setups in \S\ref{section:in_out}. First, we describe our general experimental setup.

\begin{figure}[t]
   \centering
    \begin{subfigure}[t]{0.5\textwidth}
    \vskip 0pt  %
    \includegraphics[clip,width=\textwidth]{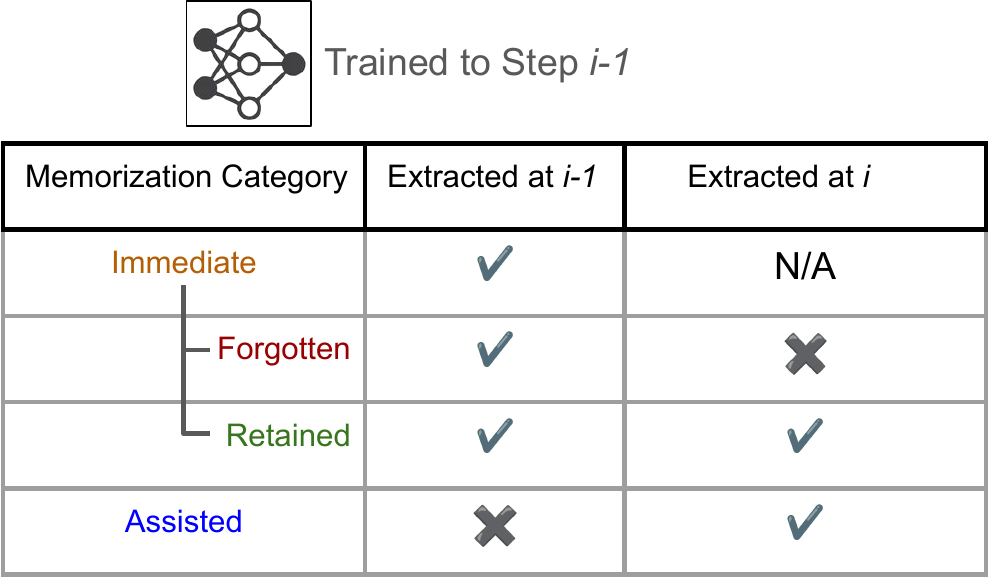}
    \end{subfigure}
  \caption{\textbf{Taxonomy of memorization for a continuous training setup.} We define \novel, \retained, \forgotten, and \assisted (described in Section~\ref{sec:taxonomy}). Note that text classified as \assisted memorization may also be forgotten or retained for steps $i+1$ onwards.}
  \label{fig:taxonomy}
\end{figure}

\paragraph{Training Setup}
We use the GPT-2-XL model~\cite{radford2019language}, which has 1.5B parameters for our primary experiments, and also experiment with Llama 3 8B~\citep{dubey2024llama3herdmodels}\footnote{Llama experiments in this paper were conducted only by parties outside of Google.} and Gemma 2B~\citep{team2024gemma}. We fine-tune these models with a linear schedule: initial and end learning rate of zero, $500$ step warmup, cooldown, and peak learning rate of $2 \times10^{-5}$. We use $1 \times10^{-2}$ weight decay and a batch size of $8$. We run experiments 5 times, sampling fresh randomness (model weights, data order, etc.) each time. 

We fine-tune these models on two datasets. First, we use a modified version of the WikiText-2 dataset~\cite{merity2016pointer} to include unique emails from the Enron dataset\footnote{\url{https://www.cs.cmu.edu/ enron/}}. We take the entire WikiText-2 dataset and insert %
\(E\) unique email addresses (herein, emails) randomly into each passage. We perform random insertions to eliminate any contextual dependency between the emails and the surrounding text, which the model could otherwise use to predict the emails more accurately. This allows us to study memorization in a worst-case scenario where no contextual cues are available (more details in Appendix~\ref{section:dataset_construction}). We concatenate all passages during training and divide them into blocks of 128 tokens.
Second, we use the Pile of Law dataset~\citep{henderson2022pile} (Appendix~\ref{sec:appendix_moreresults_continued}).
We ensure no emails were already memorized by querying the base models with the same prompts.
\citet{lee-etal-2022-deduplicating} and \citet{kandpal} found data duplication strongly increases memorization. 
In our study, all emails occur in the training corpus exactly once. %

\paragraph{Sampling}
We closely follow the methodology of~\citet{carlini-extraction,nasr2023scalable}. We focus on ``extractable memorization'' and use ten-token sequences sampled uniformly at random from Common Crawl. We randomly sample a unique set of $25,000$ different prompts for each experiment. We obtain a $256$ token output from the model for each prompt and evaluate it for successful extraction. Our method may lead to false negatives; however, this would only underestimate the PII regurgitation, and, we further believe our diverse and large prompt dataset reasonably captures the regurgitation rates. To further minimize false-negatives, where denoted we also evaluate ``discoverable'' memorization, where we prompt with the exact prefix the model was trained on (Appendix~\ref{section:hyperparams} Section~\ref{section:disc_vs_extracted}).
We use greedy decoding, or top-$k=40$ sampling when specified.

\paragraph{Defining Memorization and Extraction}
We primarily use the definition of \emph{extractable memorization} (and, where denoted, \emph{discoverable memorization} (Section~\ref{section:disc_vs_extracted})) from 
~\citet{nasr2023scalable}. Herein, we will refer to a success as an extraction, which is whenever an \emph{email} is contained both in the training dataset and a language model's generation.
Formally, let $\dataset$ be the training dataset for a language model $\lm$. Let $f$ be a chosen sampling scheme that takes an input text prompt $p$ and returns the conditional generation $s=f_\lm(p)$. An email $e^i$ is said to be extracted if $e^i \in \dataset$ and $\exists p : e^i \in f_\lm(p)$.

\paragraph{Checking for Memorized PII}
We use a regular expression to identify any emails within the generations that belong to the model's training data. Unlike previous approaches that create a pool of generations by filtering based on factors like perplexity and entropy \citep{carlini-extraction}, we evaluate all 25,000 generations for memorization.

\section{A Dynamic Lens on PII Memorization}
\label{section:continuous_training}

Production language models today consist of many training stages (pre-training, post-training, product-specific fine-tuning, etc.) and may be continually updated or refreshed with new data, e.g., to incorporate new human data using RLHF~\citep{stiennon2020learning}.
These stages may incorporate varying degrees of personal information.
This raises the question: \emph{how does memorization of sensitive data like PII evolve in this dynamical system?}

\paragraph{Continuous Training Setup.}
To study this question, we use the simplest setup that generally captures all of the above scenarios: we study memorization throughout supervised fine-tuning. We train a model by keeping the rate of emails seen constant and save checkpoints at regular intervals (for efficiency, only every 10\% of training).
Details on the dataset construction are in \S\ref{sec:prelim}.

\subsection{Categorizing Memorization Phenomena}
\label{sec:taxonomy}

Memorization analysis is typically based on \emph{only} the final model, in both academia~\citep{quantifying} and industry~\citep{team2024gemini,dubey2024llama3herdmodels,team2024gemma2}.
We now present our taxonomy for dynamic memorization analysis and use it to analyze how memorization manifests throughout continual training.

We begin by looking at the first step of training. There are but two options for any PII seen in this step: for the model to memorize it, or not. We call this type of memorization \novel, since by construction our dataset contains this email exactly once. Now, say this model were trained for another step. This new model may observe new (\novel) memorization. Beyond this, we would expect that the rest of the memorization overlaps with the prior model, which we call \retained memorization, similar to analysis in~\citet{biderman2023emergent}. Finally,~\citet{jagielski2022measuring} would tell us that we may also expect some sequences to be \forgotten.
\emph{However, we observe an additional phenomenon: \assisted memorization.} This occurs when PII not memorized at the immediate checkpoint becomes extractable later in training. We discuss this in more detail in \S~\ref{section:assisted_memorization}. Figure~\ref{fig:taxonomy} shows our complete memorization taxonomy.

\subsection{Experimental Results}
\begin{figure*}[t]

  \includegraphics[width=1.0\textwidth]{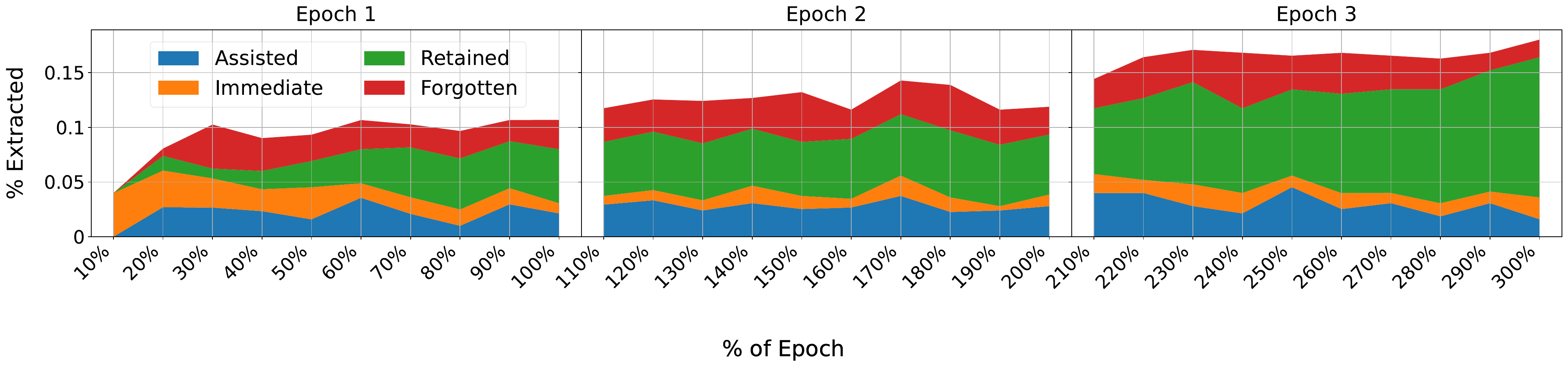}  
  
  \caption{\textbf{Tracking memorization throughout training with our taxonomy.}* The stacked bars show how many newly memorized emails are \novel, \retained, and \assisted, while red denotes forgotten emails since the last checkpoint. We see large amounts of \assisted memorization occurring later in training, underscoring that PII is not always memorized immediately. \textit{Takeaway:} memorization is more dynamic and stochastic than often assumed, with ongoing cycles of \textcolor{red}{forgotten} and newly \assisted emails. *Total number of extracted emails at each checkpoint are normalized by the number of emails seen until that checkpoint.} 
 
  \label{fig:B_greedy_normal}
\end{figure*}

Using this taxonomy of \novel, \retained, and \forgotten memorization (and \assisted memorization), we characterize all the extracted emails we observe throughout training (using the setup described above). 
Our results are shown in Figure~\ref{fig:B_greedy_normal}. We observe that there is a trend that more \novel memorization occurs near the beginning of training, whereas there is a lower rate of \novel memorization later in training.
This trend is particularly true for larger models, likely because these models memorize faster.

We also find that models are constantly \forgetting. Throughout the entirety of training (including the beginning and end), many models (see Appendix~\ref{sec:appendix_moreresults_continued} for more results on other models and datasets) exhibit a cycle of \forgetting and \novel memorization. This result sheds new light on the dynamic view of memorization: which samples are memorized by a model may be more a function of stochasticity than previously thought. The choice of which model to release may play a larger role in determining which samples are memorized, due to which samples were \forgotten or re-memorized than previously thought due to the stochasticity in data sampling.

\paragraph{Not all memorization occurs immediately.} When using our taxonomy to analyze memorizing, we observe that a significant fraction of memorization samples are not classified by these three categories. This leads to another interesting finding: a lot of memorization is \emph{not} \textcolor{orange}{immediately} memorized.
In other words, at a given step, other text that \emph{was not trained on at this step} is now extractable by the model. %

\begin{figure}[t]
  \centering
  \includegraphics[width=1.0\columnwidth]{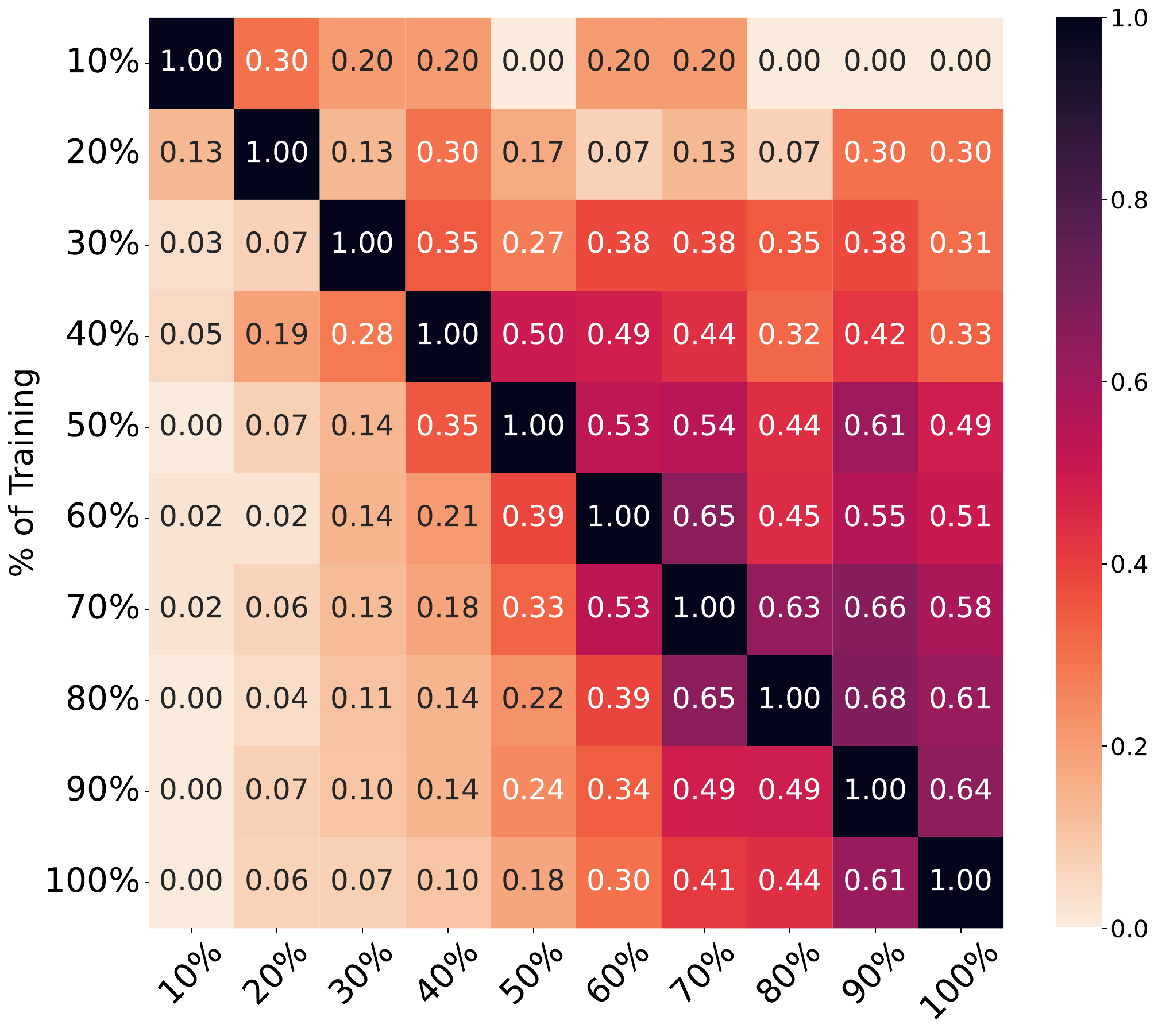}
 
  \caption{\textbf{Forgotten PII is re-extracted later.} The diagonal values \(d_{ii}\) (reference cells) represent the total extraction at each checkpoint; off-diagonal cells show the fraction of emails from the reference cell that are also memorized in the current cell. 
  \textit{Takeaway:} memorized PII can sometimes slip out of memory, only to reappear once certain overlapping tokens occur in future training steps.}
  \label{fig:rememorization}
 
\end{figure}

\paragraph{Forgetting and Re-Extraction of PII.}
Our results in Figure~\ref{fig:B_greedy_normal} show that LLMs do forget some of the previously memorized PII as training progresses. 
Prior work has shown that some examples memorized early in training may be forgotten after additional training~\citep{jagielski2022measuring}. 
Further, we also observe that some \emph{forgotten} emails get \emph{re-extracted} when there is \ngram overlap between tokens from the email and tokens in the data during further training. This phenomenon is illustrated in Figure~\ref{fig:rememorization}, which shows how previously extracted samples that the model later \emph{forgets} can reappear at subsequent checkpoints. Each cell indicates the percentage of emails extracted by both the corresponding checkpoint and the reference checkpoint (diagonal cell). Since each diagonal cell serves as its own reference, its value is always 1. 

\section{Assisted Memorization: Training on One's PII Can Reveal Another's}
\label{section:assisted_memorization}

In Figure~\ref{fig:B_greedy_normal}, we see that a large fraction of memorization is \assisted. This is especially true later in training, where we observe that more memorization is \assisted than \novel, specifically a mean rate of $0.03$ for \assisted compared to $0.01$ for \novel. This finding is not model- or data-specific, as our results in Appendix~\ref{sec:appendix_moreresults_continued} (e.g., Figures~\ref{fig:imm_vs_assi} and \ref{fig:imm_vs_assi_auc}) show similar trends.

The existence of \assisted memorization brings to light a deeper privacy concern. One may expect that data seen earlier is less vulnerable to privacy risks through a form of ``recency bias'' (implied by \forgetting effects). 
Our findings of \assisted memorization, however, show that this may not always be the case; the existence of this effect with sensitive data like PII is of particular concern because it shows that downstream training stages must be careful how they may elicit the extraction of earlier training data. The most common practical scenario for this is in the pre-training/fine-tuning setup that current LLMs undergo. Our results show that fine-tuning even on natural (non-adversarially) constructed training datasets can uncover the extraction of PII in pre-training data. Prior work~\citep{nasr2023scalable} only showed that this may be possible with adversarial constructions. Pragmatically, our results also show that privacy and memorization audits, especially when PII is of concern, should encompass all data in the training history, and not just data from the most recent training stage.

\subsection{Assisted Memorization Is Not Simply Delayed}
\label{sec:not-delayed}
Above, we found that extraction can be elicited at training steps later than where a piece of sensitive text was seen during training, in what we call assisted memorization. Here, we explore to what degree this assisted memorization is assisted by particular text in the training data, or if it was inevitable and simply delayed.

We find emails that were identified as assisted memorization at various points in training. Our aim is to re-perform training between when they were first seen and when they were later extractable by selecting entirely fresh data from the remainder of the (unseen) training dataset. Then, we can observe if only this unique set of data elicited the memorization or if any batch could.

We know when data samples were first seen from data sampling. Then, we must identify exactly when each email became extractable, as any training beyond this may lead to \forgetting. Given that we only checkpoint our models every 10\% of training, for efficiency, we do not have this a priori. 
To determine this, we use a binary search, performing an extraction test on each iteration of the search. This significantly reduces the overhead as the extraction test is expensive (recall we prompt the model thousands of times as described in \S\ref{sec:prelim}). 

Overall, we run this procedure on four unique emails and with seven trials each. We find that emails became extractable in only $35.7\%\pm15.9$ of them on average. While this refutes the idea that there may be a single unique set of data that leads to assisted memorization, this shows that most sets of data do not lead to it. Next, we explore what characteristics the successful trials share.

\subsection{Assisted Memorization Is Triggered by Training on Specific \ngrams}
\label{section:sub_assisted}
Our analysis here is inspired by~\citet{lee-etal-2022-deduplicating}, who show that data repetitions (duplication) heavily influence memorization of text. While our data setup in \S\ref{sec:prelim} has no exact duplicates of these emails, there can still be overlaps of important \ngrams.

\paragraph{Causally Removing \ngrams.}
To study this, we perform a causal intervention whereby we remove all training sequences that have high \ngram overlap with emails identified as assisted memorization. We use a similar setup to the previous \S\ref{sec:not-delayed} except that we notably remove any text that overlaps with the assisted memorized emails. For each trial of this experiment, we select a different checkpoint $\lm_i$ throughout our continuous fine-tuning setup; let $\dataset_i$ be the set of training sequences used to train $\lm_i$ from $\lm_{i-1}$. We take all emails identified as assisted memorization on $\lm_i$; for each, we construct a simple regex-based filter that checks for names in the email address based on common email formatting patterns (e.g., name@gmail.com or firstname.lastname@gmail.com). We use these regex filters to remove any text in $\dataset_i$ and then retrain $\lm_i$ from $\lm_{i-1}$ on this new dataset. 

Across all 30 checkpoints and 5 seeds, we find a total of 177 emails that were assisted memorized. After intervening to remove overlapping \ngrams from batch $D_{i}$, all but 10 of these assisted memorized emails were no longer memorized.

\paragraph{Features Associated with Memorization} Next, we ask: when multiple emails share a \texttt{firstname}, why might a particular email with a different \texttt{lastname} get assisted memorized over another? For example, why might \texttt{john.mccarthy@gmail.com} be memorized over \texttt{john.williams@gmail.com}. We train a simple logistic regression model on features capturing \ngrams overlaps, last-name counts, and domain counts for all assisted memorized emails (positives) and those not memorized (negatives). More details are in Appendix~\ref{sec:additional-assisted}.

Our logistic regression model is trained to predict assisted memorized emails from a dataset consisting of these emails labeled as positive, and other emails sharing the same \texttt{firstname} but a different \texttt{lastname} as negatives. We use a standard 5-way cross validation setup with 10 trials. Full details are in Appendix~\ref{sec:additional-assisted}. The model achieves a precision of 0.937 and recall of 0.874 indicating high success.

\begin{figure}[t]
  \centering
  \includegraphics[width=1.0\columnwidth]{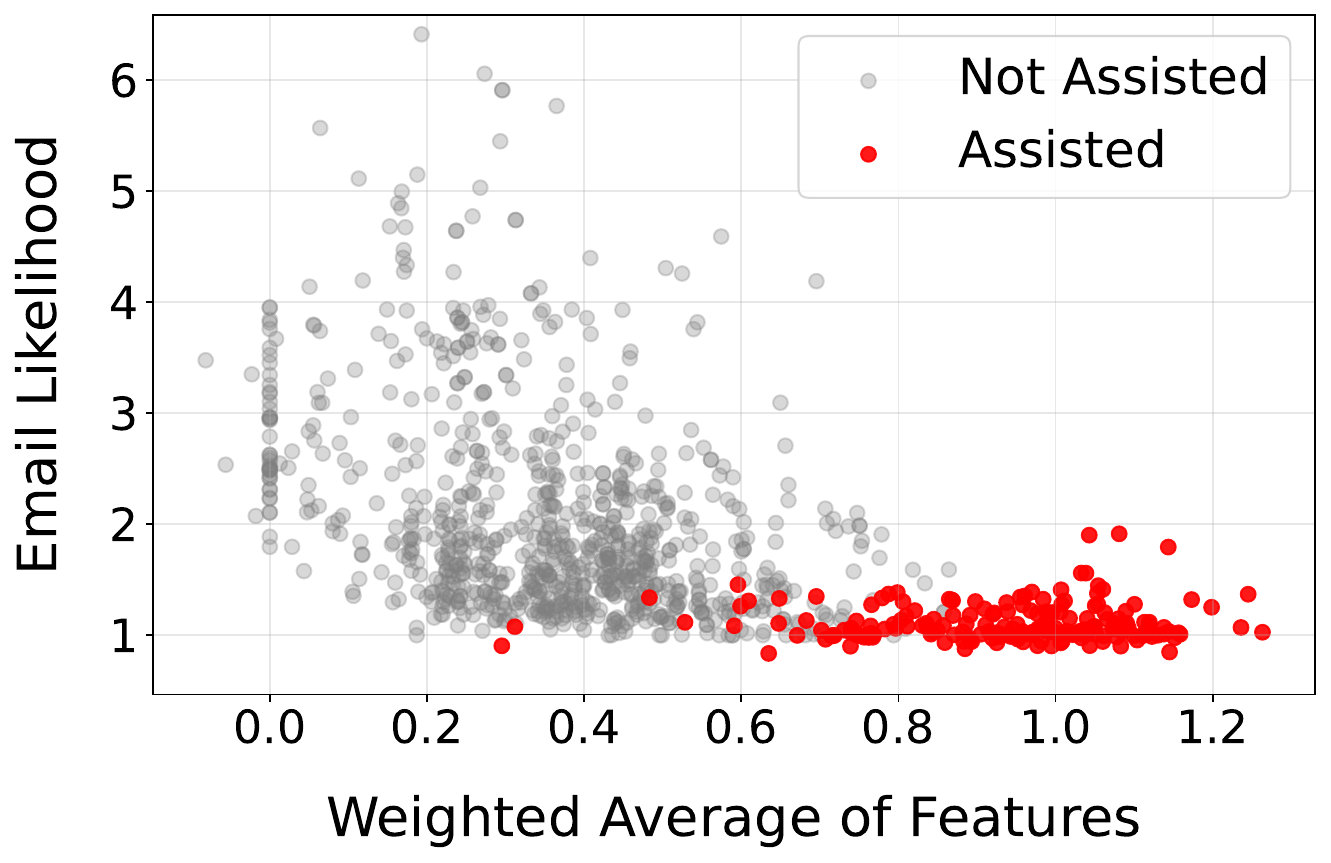}
 
  \caption{\textbf{Overlap features predict which emails are assisted memorized.} We plot a logistic-regression score (x-axis) vs.\ conditional likelihood (y-axis). Emails that become assisted memorized (red) exhibit higher \ngrams overlap (i.e.\ higher model score), whereas those not memorized (grey) have lower overlap. 
  \textit{Takeaway:} overlapping \ngrams in future training data strongly drive which PII is triggered to appear in the model’s output.}
 
  \label{fig:assisted_scatter}
\end{figure}

In Figure \ref{fig:assisted_scatter}, we visualized the logistic regression model's score against the email likelihood from $\lm$, computed against the successful prompt that led to extraction. This shows that \assisted memorization emails tend to be well classified from these simple features.
We observe that \ngram statistics were the most important feature, further supporting our conclusions above (see Table~\ref{tab:weights} of Appendix~\ref{sec:additional-assisted} where we report the feature weights).

\section{Do PII Opt-ins/Opt-outs Impact Extraction?}
\label{section:in_out}
\subsection{Contributing More Data via Opt-ins}
\label{section:opt-in}
If many new users opt-in to contribute data to a model, then the model owner may want to incorporate new information (and sometimes, new PII) into the finetuning pipeline. 
One of the simplest ways to do this is by adding the new PII to existing training data and re-finetuning the model from scratch. From our results in \S\ref{section:assisted_memorization}, we know that continuing to train a model on additional PII could lead to increased extractability of previously unextracted PII. 
In this section, we study how retraining with additional PII changes the extractability of prior data.\looseness=-1

\paragraph{Setup} 
To mimic the above scenario, we design a \textbf{Retraining Experiment} where we add more emails to the existing dataset and re-finetune the model on the updated dataset. %
We write \(D_{x\%}\) as the finetuning dataset containing $x\%$ of the emails from the global set of emails $X$. We construct 10 different finetuning datasets containing increasing amounts of emails: $D_{10\%}, D_{20\%}, \cdots, D_{100\%}$. In $D_{x\%}$, we include $x\%$ of the global pool of emails $X$, such that, if $a<b$, all emails that are found in $D_{a\%}$ are also found in $D_{b\%}$. Before constructing these datasets, we randomly shuffle the emails in $X$ to ensure a uniform distribution of emails in each dataset.\looseness=-1

Next, we train ten distinct models \(M_{1}\) to \(M_{10}\), where $M_i$ is trained on $D_{10i\%}$ for three epochs, following the same training setup described in \S\ref{sec:prelim}. We highlight that the only change between these models is the additional emails. Otherwise, we use the same training process and the same prompts for all models when decoding.

\paragraph{Adding More PII Increases Extraction of Existing PII.}
We report the results of our experiment in Figure~\ref{fig:add_topk_greedy_e3}, for models finetuned for three epochs (more results in Appendix~\ref{section:addition_moreresults_retrained}). We highlight two major findings.

First, we find that the number of extracted emails increases substantially with the amount of PII contained in the model's fine-tuning set. This can be seen on the diagonals of Figure~\ref{fig:add_topk_greedy_e3}, which show the total amount of PII extracted from the relevant model. For top-$k$ sampling, we see that 283 emails are extracted from $M_{10}$, compared to only 57 at $M_{5}$, which was trained on half as many emails---the increase in extraction from top-$k$ sampling is superlinear in the fraction of emails included in the model's finetuning set. The increase is still substantial, but not superlinear, for greedy sampling.

Our second and main finding is that the inclusion of more PII leads to \emph{existing} PII being at higher risk of extraction from top-$k$ sampling. This can be seen from the general positive trend in extracted emails for each dataset $D_{x\%}$ along the $x$ axis. To validate this result, we run a binomial hypothesis test, for whether top-$k$ sampling extracts more emails from $D_{i\%}$ when run on $M_{j}$ ($j > i$) than when run on $M_{i}$. With 45 such comparisons, 41 show more extraction for models which see more emails ($p < 10^{-8}$, and $p < 10^{-4}$ for 1 and 2 epochs).

\begin{figure}[t]
  \centering
  \includegraphics[width=1\columnwidth]{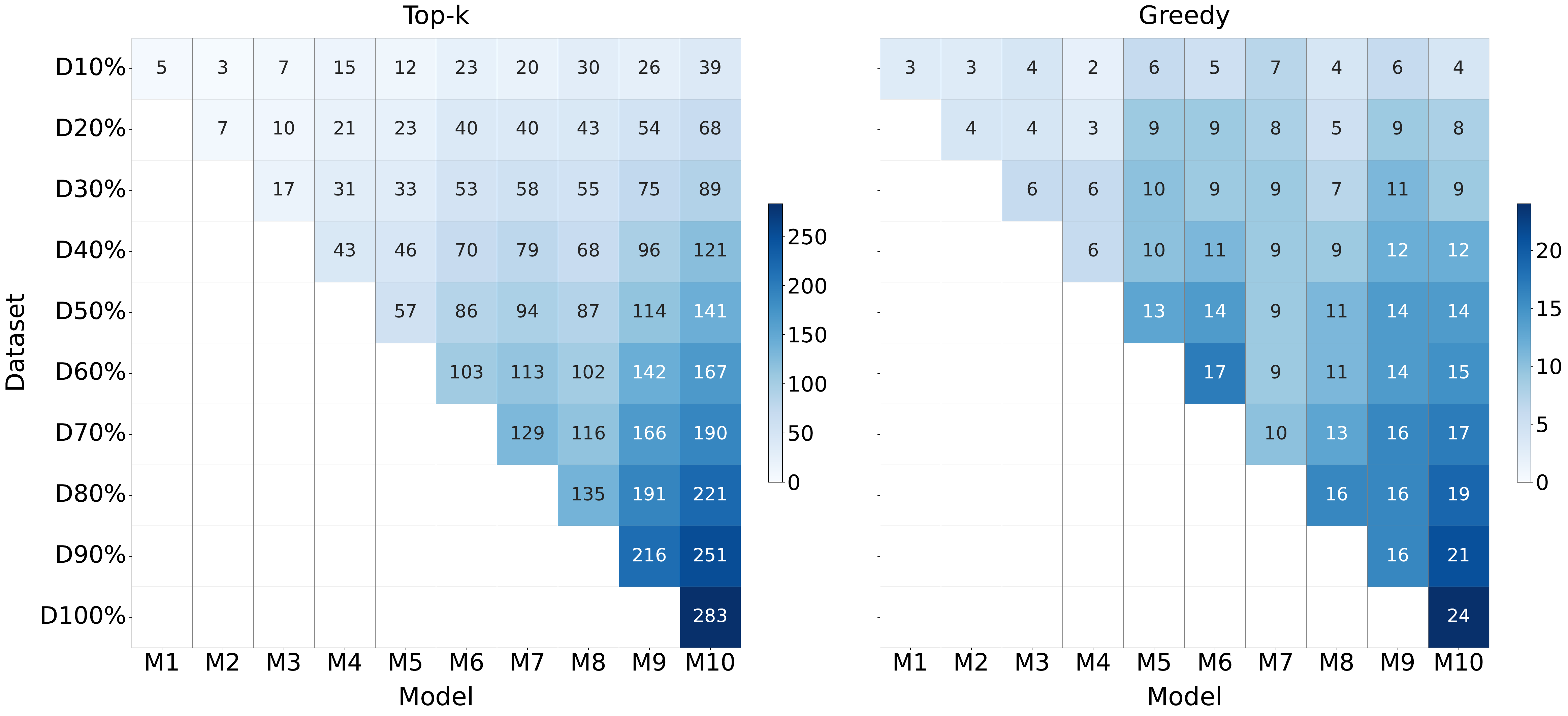}
 
  \caption{\textbf{Adding more PII leads to more extraction.} Each row corresponds to a dataset $D_{x\%}$, and each column corresponds to the model $M_j$ trained with $j\times 10\%$ of the emails. The values show how many emails in $D_{x\%}$ are extracted by $M_j$. 
  \textit{Takeaway:} introducing new PII during re-finetuning (moving along the x-axis) also increases the extraction \emph{of old PII} that was already present in the training set. This effect can increase extraction by a factor of over $7\times$ in our settings, as seen in the extraction of emails in $D_{10\%}$ from $M_{10}$.}
  \label{fig:add_topk_greedy_e3}
\end{figure}

\begin{figure}[t]
  \centering

  \includegraphics[width=1.0\columnwidth]{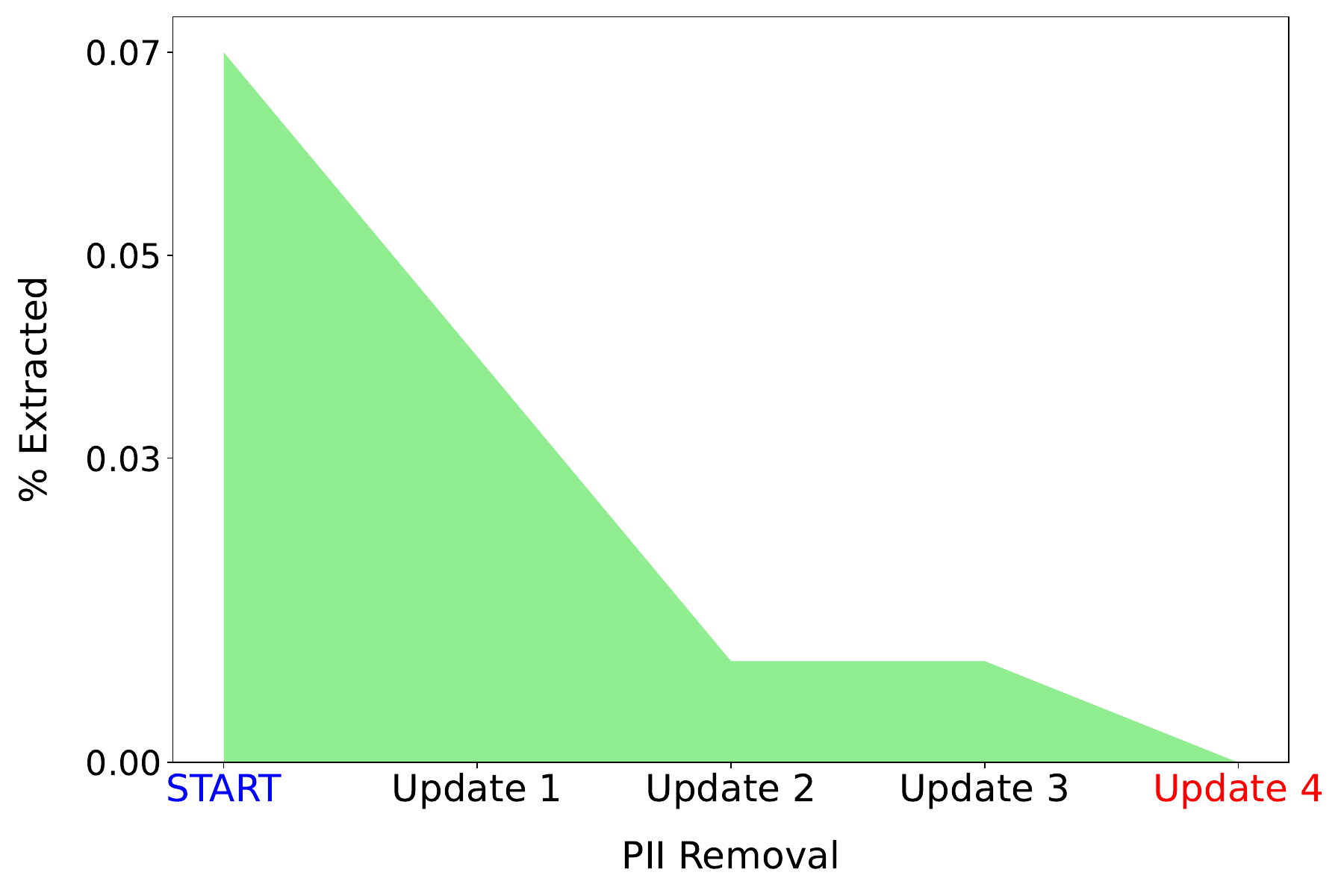}
 
  \caption{Removing extracted PII from the training data and retraining can lead to new memorized PII. 
  After four removal-and-retrain cycles (\textcolor{red}{Update~1}--\textcolor{red}{4}), 
  no additional PII is extracted under the same 25k prompts and greedy decoding. 
  \textcolor{blue}{START} denotes the original model.}
 
  \label{fig:onion}
\end{figure}

\subsection{Protecting PII via Opt Outs}
\label{section:onion}

As data opt-outs are becoming increasingly common on the web~\citep{linkedinoptout}, %
we first study how removing a user's PII from the training data can inadvertently trigger the extraction of additional PII.
We then investigate factors that correlate to PII becoming extractable once similar PII is removed.\looseness=-1 %

\paragraph{Setup}
\label{subsec:ppi_setup}
We study the simplest unlearning technique, often referred to as \emph{exact machine unlearning}~\citep{bourtoule2021machine}: removing all relevant PII from the dataset and retraining, or as here re-fine-tuning, the model.
This may be triggered if users submit an opt-out request.
Since retraining after each request is expensive, %
model owners may collect and process these requests in batches.\looseness=-1

Following a protocol similar to~\citet{onioneffect}, our experimental procedure is:
     \textbf{(1) Extraction:} Prompt the current model \(\mathcal{M}\) with 25,000 fixed prompts and sample using greedy decoding to identify memorized emails. Let \(E\) be the set of extracted emails.
     \textbf{(2) Removal:} Remove \(E\) from \(D\) and re-finetune the base model on \(\,D \setminus E\), producing a new model \(\hat{\mathcal{M}}\).
     \textbf{(3) Repeat:} Prompt \(\hat{\mathcal{M}}\) again with the same prompts, discovering any newly memorized emails \(\hat{E}\).
We iterate until no more emails are extracted using this fixed set of prompts and decoding strategy.

\begin{figure}[t] 
\centering
\includegraphics[width=1.0\columnwidth]{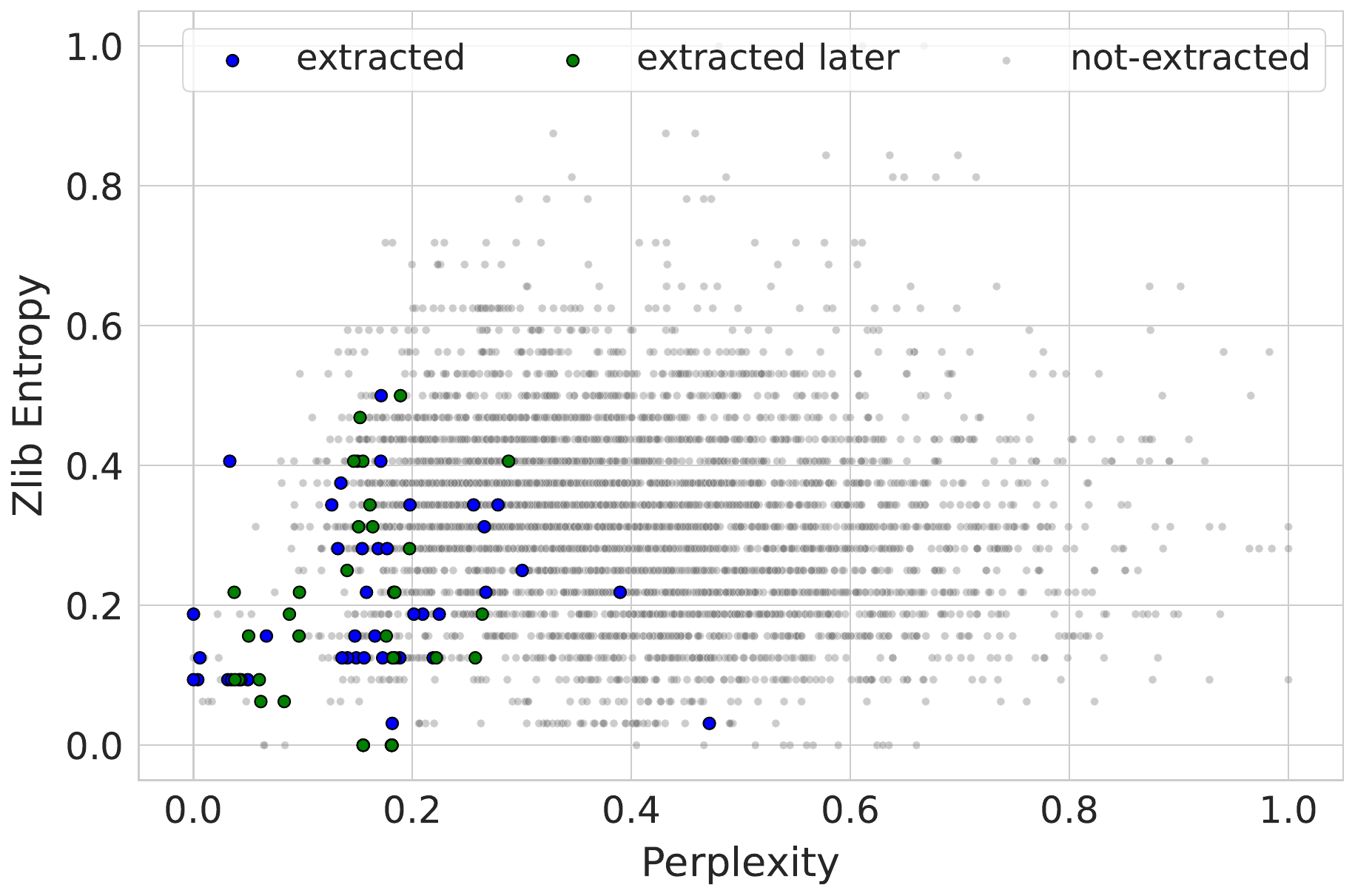} 

\caption{{Perplexity and zlib entropy of memorized emails.} Emails extracted in the initial model (blue) and emails extracted in later re-finetuned models %
(green) have lower perplexities than emails that were never extracted by any model (grey). This clustering suggests that the newly-extracted (green) emails were near the threshold of memorization from the outset.\
} \label{fig:entropyvsppl}

\end{figure}

\paragraph{Protecting One Person's PII May Leak Another's}

As mentioned above, in each iteration, we (1) prompt the current model \(\mathcal{M}\) (trained on dataset \(D\)) with 25,000 fixed prompts, (2) remove any newly discovered memorized emails \(E\) from \(D\), and (3) re-finetune the base model on \(D \setminus E\). 
Figure~\ref{fig:onion} illustrates four such rounds (\textcolor{blue}{START} through \textcolor{red}{Update~4}). 
While the first update successfully removes the previously identified emails from the set of extracted PII, it simultaneously extracts a \emph{new} set of emails. 
By \textcolor{red}{Update~4}, no additional emails are discovered under these prompts and greedy decoding, although changing prompts or sampling strategies could still reveal further memorization. 
 Our results confirm that this \emph{layered memorization}---called the Onion Effect by prior work on image classifiers~\cite{onioneffect}---extends to language models: removing one layer of memorized PII exposes a second layer, and so forth.

\paragraph{Removing Random Emails.}
We next conduct a similar experiment but remove a random subset of emails instead of the ones that are discovered through extraction. Specifically, we sample 10\% of the total emails in \(D\) uniformly at random and call this set \(E\). We then fine-tune a new model \(\hat{\mathcal{M}}\) on \(D \setminus E\). Prompting \(\hat{\mathcal{M}}\) with the same 25,000 prompts and sampling with greedy decoding yields a new set of extracted emails \(\hat{E}\). 
Thus, \emph{randomly removing} data can similarly expose new PII, underscoring how unlearning updates can inadvertently introduce new privacy risks.

\paragraph{Controlling for Randomness During Training.} 
A natural question is whether any newly extracted emails simply result from any randomness when retraining a new model. 
For instance, models trained with the same data order, same parameter initialization, and same hyperparameters could
still differ during inference as GPU operations are non-deterministic~\citep{highfidelityextraction}. We want to ensure that new extractions are solely the result of removing particular emails. To this end, we train five such new models and extract emails by feeding the exact same prompts that we give to our original model (\(\mathcal{M}\)) and the models trained after removing extracted and randomly sampled emails (\(\hat{\mathcal{M}}\)). 
We sample all three sets of models with greedy decoding and compare which emails were extracted. %
Across all five trials and for both types of removals (removing extracted emails and removing them randomly), the models re-finetuned-after-removal reveal strictly more \emph{unique} PII than these fresh counterparts. Hence, the effect is not merely a product of random training fluctuations but rather an outcome of selectively removing data from \(D\).\looseness=-1

\paragraph{PII on the Verge of Memorization Surfaces After Others Are Removed}
\label{subsec:almost_extracted_pii}
Because we use a fixed set of prompts and greedy decoding, we hypothesize that newly extracted emails in each unlearning round were already \emph{close} to being memorized under the original model. In other words, these emails were initially ``hidden'' behind a first layer of memorized PII. Once the first layer of emails is removed, these nearly extractable emails become more vulnerable. %

To investigate this, we compare the perplexity of the initial model on three categories of emails: 
\textit{(i)} those extracted in the initial model, 
\textit{(ii)} those that are extracted in subsequent rounds of removal and refinetuning %
and 
\textit{(iii)} those never extracted by any model. 
We also measure their zlib entropy, a compression-based proxy for memorization~\citep{carlini-extraction, recite, zlib}. 
As shown in Figure~\ref{fig:entropyvsppl}, %
newly-extracted emails (green) cluster with those initially extracted (blue), indicating that both groups have lower perplexity compared to never-extracted emails (grey). This supports our hypothesis: once one layer of extracted PII is removed from the training set, the next-likeliest set of emails crosses the threshold into extraction. 
Iterating this process eventually exhausts these ``hidden layers,'' although more sophisticated prompts or sampling strategies could still uncover additional memorization.

\section{Conclusion}
We study how the actions of continually training on more data, re-training with new data, or re-training after removing data can have ripple effects for privacy. In particular, we propose the phenomenon of \Assisted Memorization where examples that aren't extracted at existing checkpoints can get extracted later. This could create a false impression of privacy for examples that don't get extracted at a particular checkpoint, as training further on similar-appearing examples could lead to their extraction. We also find that including more PII in the training data can degrade privacy of existing PII by putting them at a higher risk of extraction. Furthermore, removing particular PII examples from training data could cause other examples to be extracted. This underscores the need for more holistic audits for memorization, where examples that aren't extracted at a particular timepoint are also evaluated for any potential risks.

\section*{Limitations}
In this study, we use emails as an example of PII because they are a common form of personal information and can be readily studied using publicly available datasets, e.g., the Enron corpus. We do not examine other forms of PII, such as credit card numbers or mailing addresses, partly because they are not publicly available. However, analyzing these types of PII is important to determine whether certain categories are more vulnerable to the memorization risks identified here. We believe that our methods will generalize to other forms of PII with minor adjustments. We also observe a phenomenon akin to \emph{onion memorization}~\citep{onioneffect}, where removing particular emails from the dataset and retraining the model (\emph{exact unlearning}~\cite{sisa}) can cause new emails to be extracted. A promising direction is to investigate whether this effect persists under \emph{approximate} unlearning techniques (e.g.,~\citep{hayes2024inexact}), where the model is not fully retrained from scratch. Furthermore, our focus here is solely on extraction risks for training-data emails, but other generated or partially memorized emails could also pose privacy concerns---particularly if they can serve as keys to uncover additional information about specific individuals.

\section*{Ethics Statement}

We rely on the publicly available Enron Corpus to create our fine-tuning datasets, acknowledging that some of its contents may include sensitive or personally identifiable information. To mitigate privacy risks, we follow standard diligence practices for data handling. While no additional raw text or private details are disclosed beyond those already publicly released, we analyze memorization specifically to highlight risks inherent in large language models, rather than to reveal more personal data. Our experiments use established public models and datasets (GPT-2 family, Gemma 2B, Llama 3 8B, Wikitext, and Pile of Law) to facilitate reproducibility while maintaining responsible data practices. We align our work with accepted norms for ethical use of legacy datasets like Enron and emphasize the importance of privacy-preserving training and unlearning techniques for future systems.  

\bibliography{anthology, custom}

\clearpage
\appendix
\section{More Details on Dataset Construction}
\label{section:dataset_construction} 
While we insert emails into each message at random positions to study the worst-case scenario for memorization, we also want to make sure that the utility of our fine-tuned model is not degraded. To this end, we compare the perplexity values of the original and fine-tuned models on a held-out WikiText-2, as well as a new WikiText-103~\citep{wikitext} test dataset. We compute perplexity values using a sliding window of 1024 tokens (context window of GPT-2 XL). The perplexity of the base GPT-2 XL model on the WikiText-2 test set was 15.20, while that of the fine-tuned model was 11.35. The perplexity of the base model on the WikiText-103 set was 16.49, and the fine-tuned model had a perplexity value of 13.23. These values indicate that the utility of our model is not degraded post-fine-tuning.

\section{Hyperparameters that Influence PII Extraction}
\label{section:hyperparams}
\subsection{Greedy vs. Top-\textit{k} Sampling}
\label{section:greedyvstopk}

Model owners can employ either deterministic decoding such as greedy or stochastic sampling methods (such as top-\textit{k} \cite{fan-etal-2018-hierarchical} or top-\textit{p} \cite{top-p}) to improve the quality of the generated text. Several commercial APIs providing text-generation access to models such as ChatGPT\footnote{\url{https://platform.openai.com/docs/guides/text-generation}}, Gemini\footnote{\url{https://ai.google.dev/gemini-api/docs/text-generation?lang=python}}, and Claude\footnote{\url{https://docs.anthropic.com/en/api/complete}} use a combination of \texttt{top-\textit{k}} and \texttt{top-p} parameters to generate text and extraction rates vary across sampling schemes~\citep{hayes2024measuringmemorizationprobabilisticdiscoverable}. This makes it essential to study how PII extraction varies across different sampling methods. We find that we can extract significantly more PII using top-\textit{k} sampling than greedy decoding.  

We draw the following comparisons: (1) The ratio of total emails extracted using top-\emph{k} sampling compared to greedy decoding; (2) Total emails extracted using a fixed set of 25,000 prompts for both sampling methods; and (3) Total emails generated by both sampling methods when conditioned on same 25,000 prompts.

\begin{figure*}[t]
  \includegraphics[width=\textwidth]{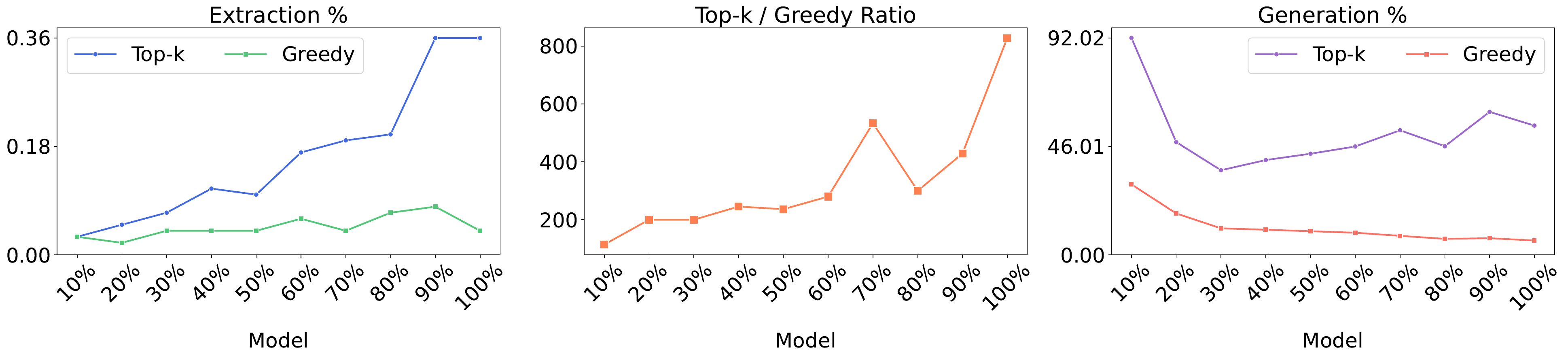}  
  \caption {(Left) We can extract significantly more emails with top-\emph{k} than with greedy decoding using the same set of prompts. (Middle) We can extract up to 800 times more emails using top-\emph{k}. (Right) top-\emph{k} generates more emails than greedy for the same amount of emails seen during training. The x-axis denotes a separate model obtained after adding an additional 10\% of total emails in the training data.}
  \label{fig:A_topkvsgreedy}
\end{figure*}

It can be seen in Figure \ref{fig:A_topkvsgreedy} that top-\textit{k} can extract emails over 800 times higher than greedy decoding. Top-\textit{k} also consistently generates more unique emails than greedy. Model owners might employ top-\textit{k} sampling as it produces more diverse and higher-quality text compared to greedy. However, this approach may pose privacy risks, such as increased memorization and leakage of personal information.

\subsection{Prompting}
\label{section:disc_vs_extracted}
We fine-tune our model on full WikiText-2 with Enron emails in it for 20 epochs and prompt after every epoch in the following manner: (1) \textbf{Extractable Prompting} using random ten-token prompts sampled from Common Crawl (as mentioned in \S~\ref{sec:prelim}), and (2) \textbf{Discoverable Prompting} where we prompt with prefixes that occur before an email in the training data (using the definition of \emph{discoverable memorization} from \citet{nasr2023scalable})

\begin{figure*}[t]
 \centering
  \includegraphics[width=0.6\textwidth]{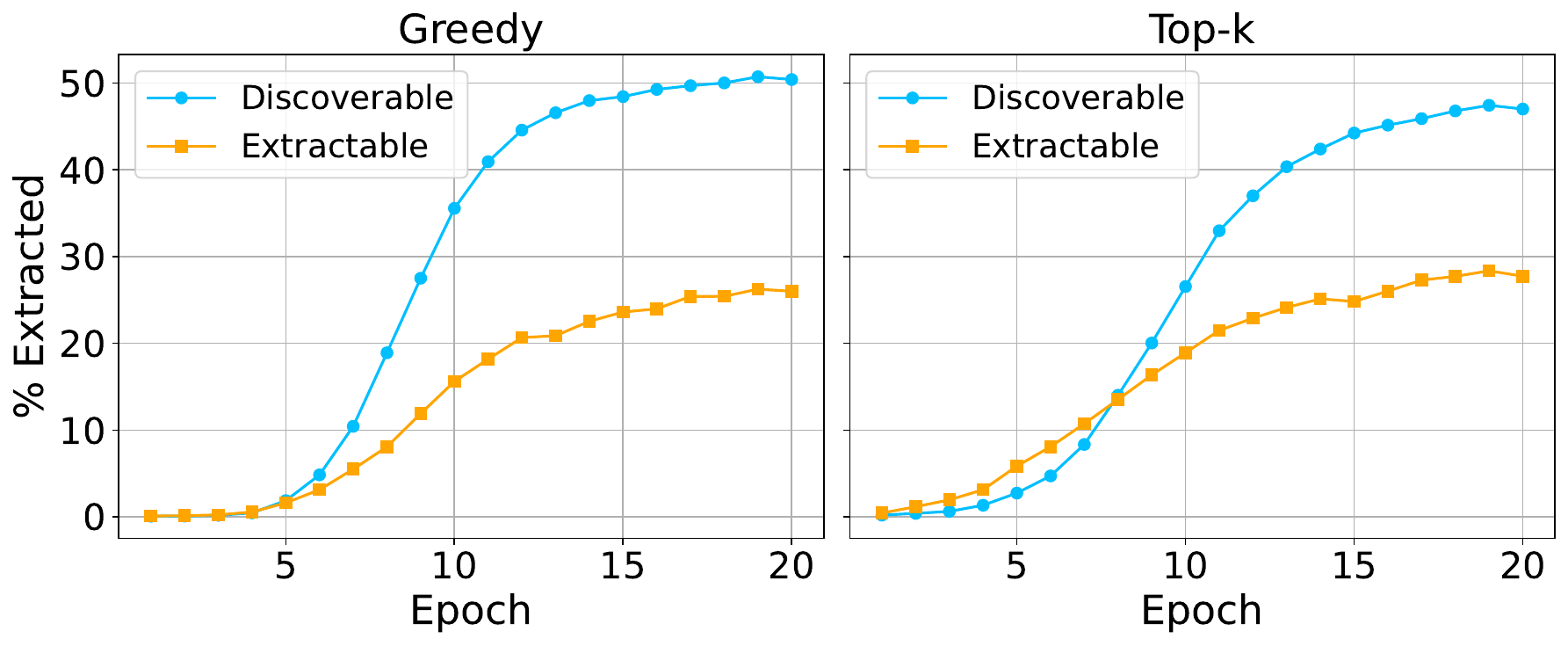}  
  \caption {Comparing extractable memorization with discoverable memorization over 20 epochs.}
  \label{fig:disc_vs_ext}
\end{figure*}

As observed in Figure~\ref{fig:disc_vs_ext}, we find that for both greedy decoding and top-\emph{k} sampling, extractable memorization is more than discoverable memorization in the initial epochs. However, discoverable memorization starts increasing significantly after epoch 5 for greedy decoding and epoch 7 for top-\emph{k} sampling. By the end of the 20th epoch, discoverable memorization is over 92\% more than extractable memorization.

\section{More Results on PII Memorization in Continuous Training.}
\label{sec:appendix_moreresults_continued}
\textbf{More results from \S~\ref{section:continuous_training}}: We fine-tune various models on two datasets—Wikitext and the Pile of Law—and show that our findings are generalizable. We only use greedy decoding for sampling from these models. 

\begin{figure*}[t]
  \includegraphics[width=\textwidth]{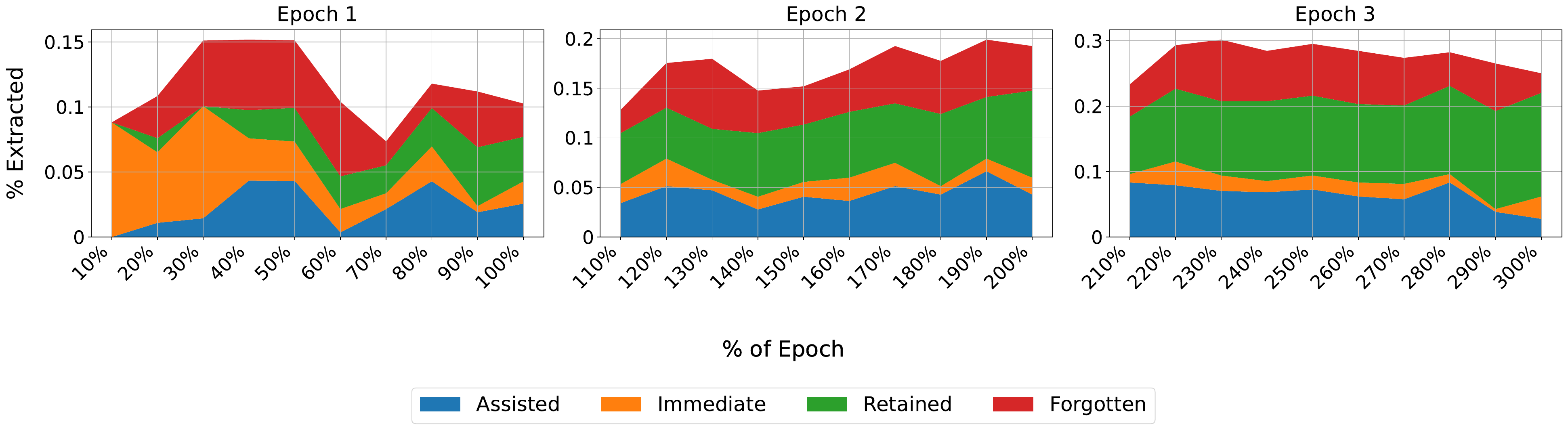}  
  \caption {Different memorization categories during continuous training for GPT-2 XL trained on the Pile of Law + Enron emails.}
  \label{fig:B_pol}
\end{figure*}

\paragraph{GPT-2 XL trained on the Pile of Law dataset:} Figure \ref{fig:B_pol} shows that our results are generalizable also on the Pile of Law dataset \citep{henderson2022pile}. We extract the \texttt{congressional\_hearings} instance from the dataset and insert enron emails in it according to our setup in \S~\ref{sec:prelim} while keeping the total number of tokens in the dataset the same as our original Wikitext dataset.

\begin{figure*}[t]
  \includegraphics[width=\textwidth]{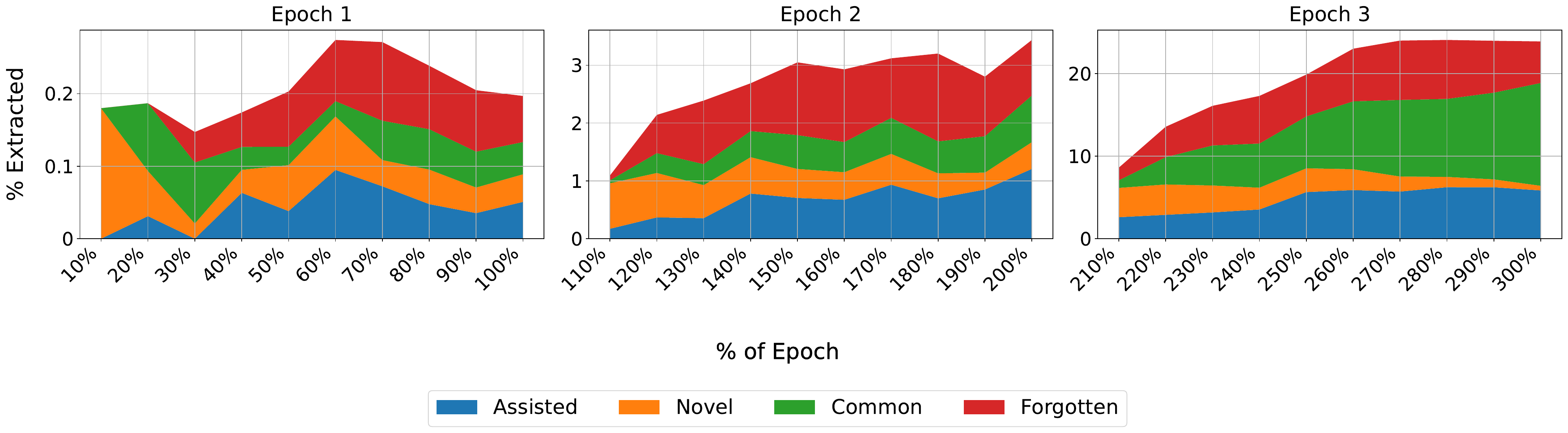}  
  \caption {Different memorization categories during continuous training for Llama3 8B trained on WikiText + Enron emails.}
  \label{fig:B_llama3}
\end{figure*}

\begin{figure*}[t]
  \includegraphics[width=\textwidth]{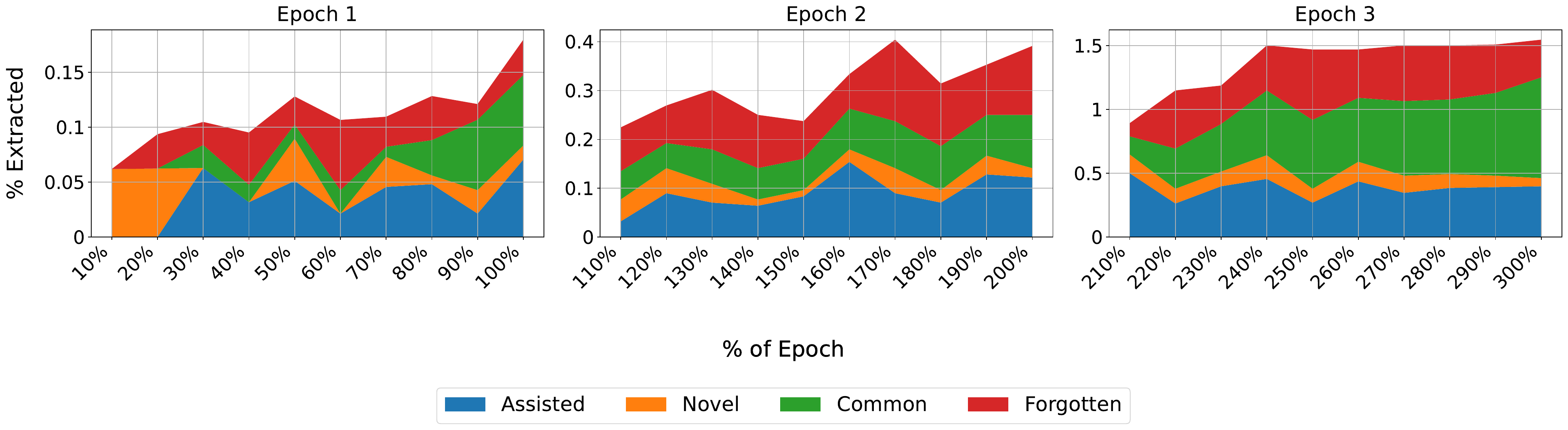}  
  \caption {Different memorization categories during continuous training for Gemma 2B trained on WikiText + Enron emails.}
  \label{fig:B_gemma}
\end{figure*}

\paragraph{Llama3 8B and Gemma 2B models trained on our original dataset (WikiText + Enron emails):} Our results generalize to the current state-of-the-art models, including Llama3 with 8B parameters (Figure \ref{fig:B_llama3}) and Gemma 2B base model \citep{team2024gemma} (Figure \ref{fig:B_gemma}). 

\begin{figure*}[t]
  \includegraphics[width=\textwidth]{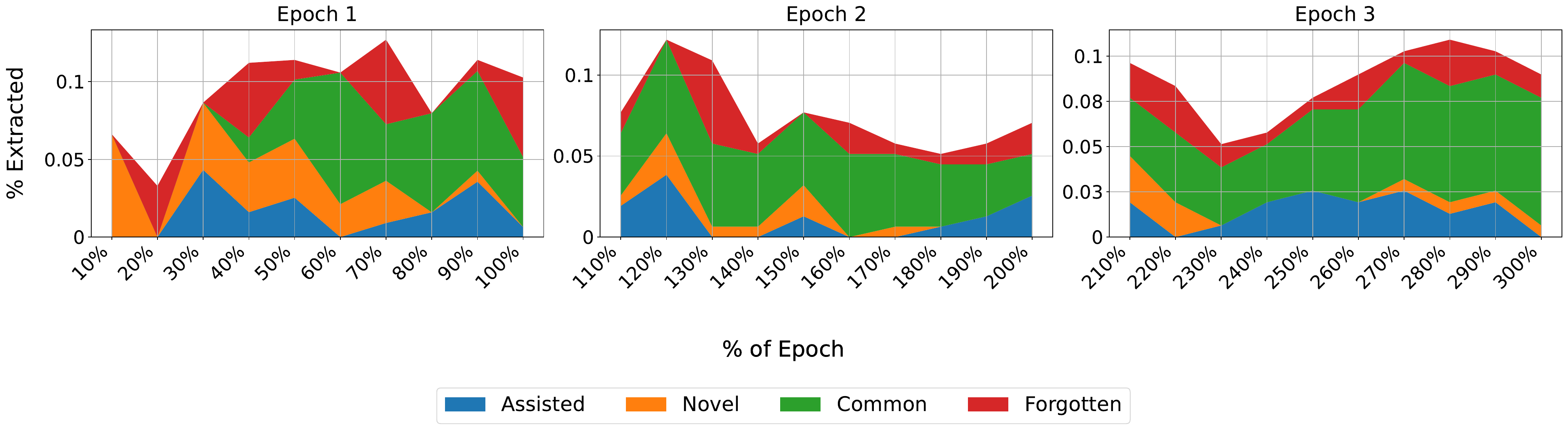}  
  \caption {Different memorization categories during continuous training for GPT-2 Large trained on WikiText + Enron emails.}
  \label{fig:B_large}
\end{figure*}

\begin{figure*}[t]
  \includegraphics[width=\textwidth]{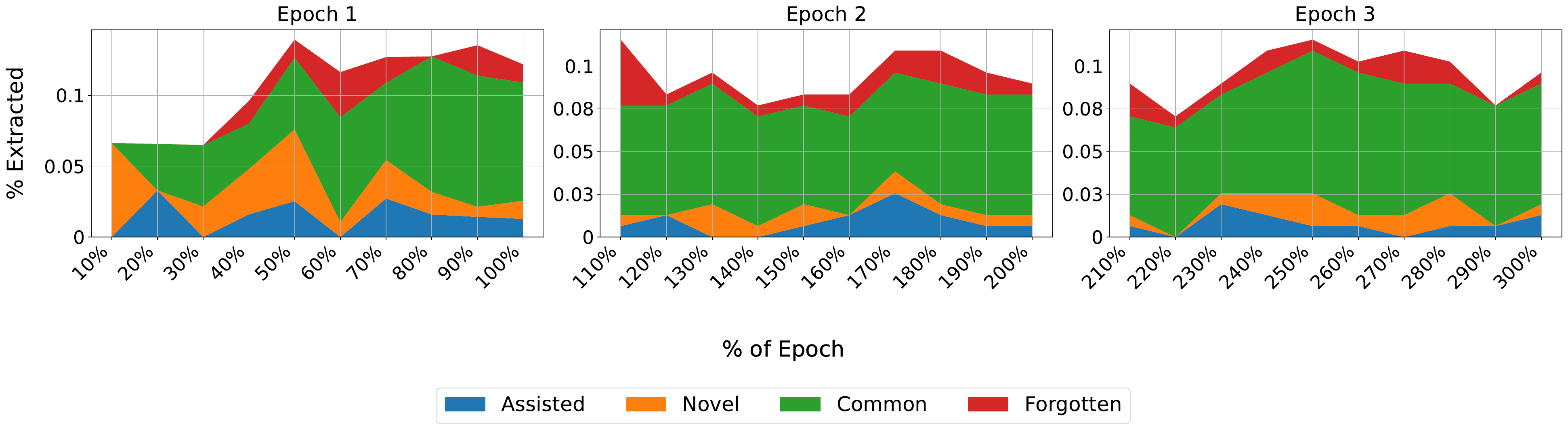}  
  \caption{Different memorization categories during continuous training for GPT-2 Medium trained on WikiText + Enron emails.}
  \label{fig:B_medium}
\end{figure*}

\begin{figure*}[t]
  \includegraphics[width=\textwidth]{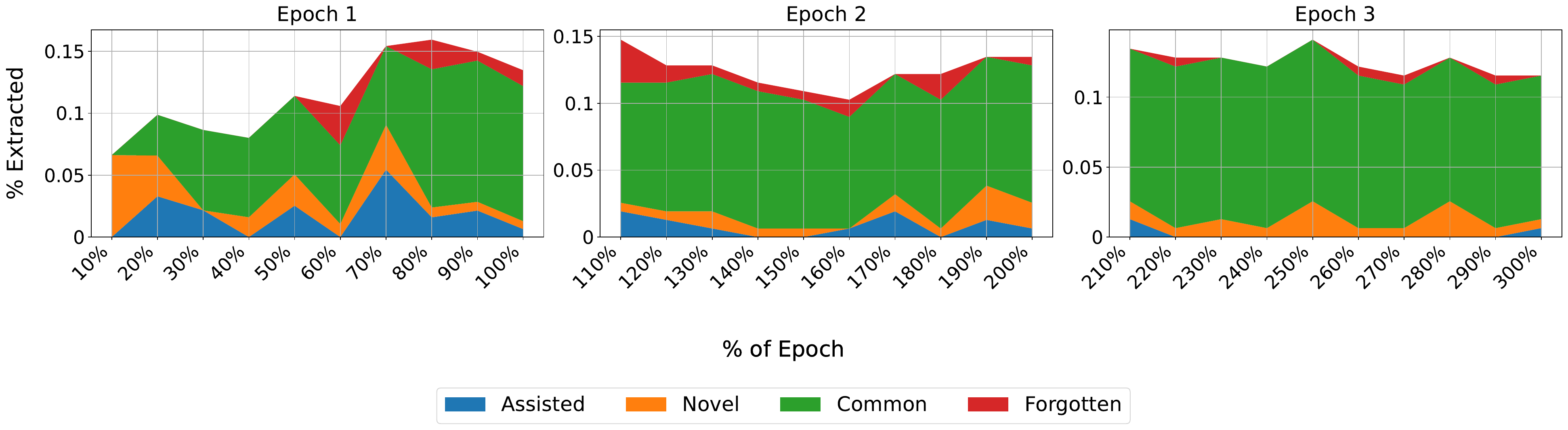}  
  \caption {Different memorization categories during continuous training for GPT-2 Small trained on WikiText + Enron emails.}
  \label{fig:B_small}
\end{figure*}

\paragraph{GPT-2 Large 774M, Medium 355M, and Small 124M models trained on our original dataset (WikiText + Enron emails):} We also train the remaining members from the GPT-2 model family: Large (Figure \ref{fig:B_large}), Medium (Figure \ref{fig:B_medium}), and Small (Figure \ref{fig:B_small}). We observe that assisted memorization becomes less prominent in smaller models. 

\begin{figure*}[t]
  \includegraphics[width=\textwidth]{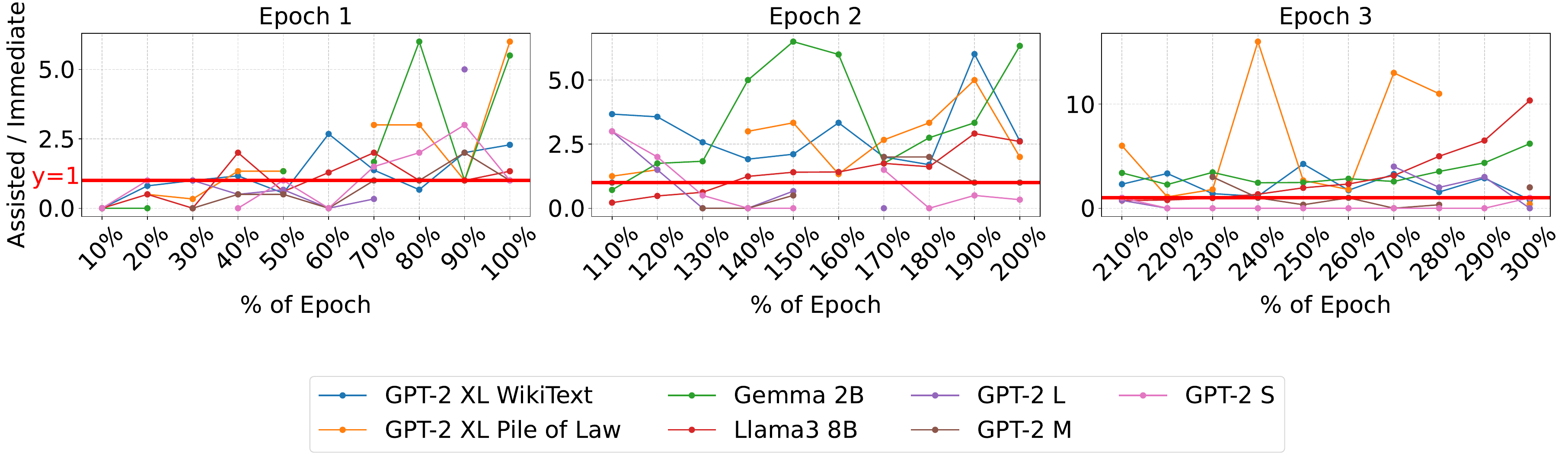}  
  \caption {Ratio of the rate of \assisted to \novel memorization. We observe that a large fraction of memorization in multiple models is \assisted. Note that some models at specific checkpoints had no immediate memorization.}
  \label{fig:imm_vs_assi}
\end{figure*}

\begin{figure*}[t]
 \centering
  \includegraphics[width=0.6\textwidth]{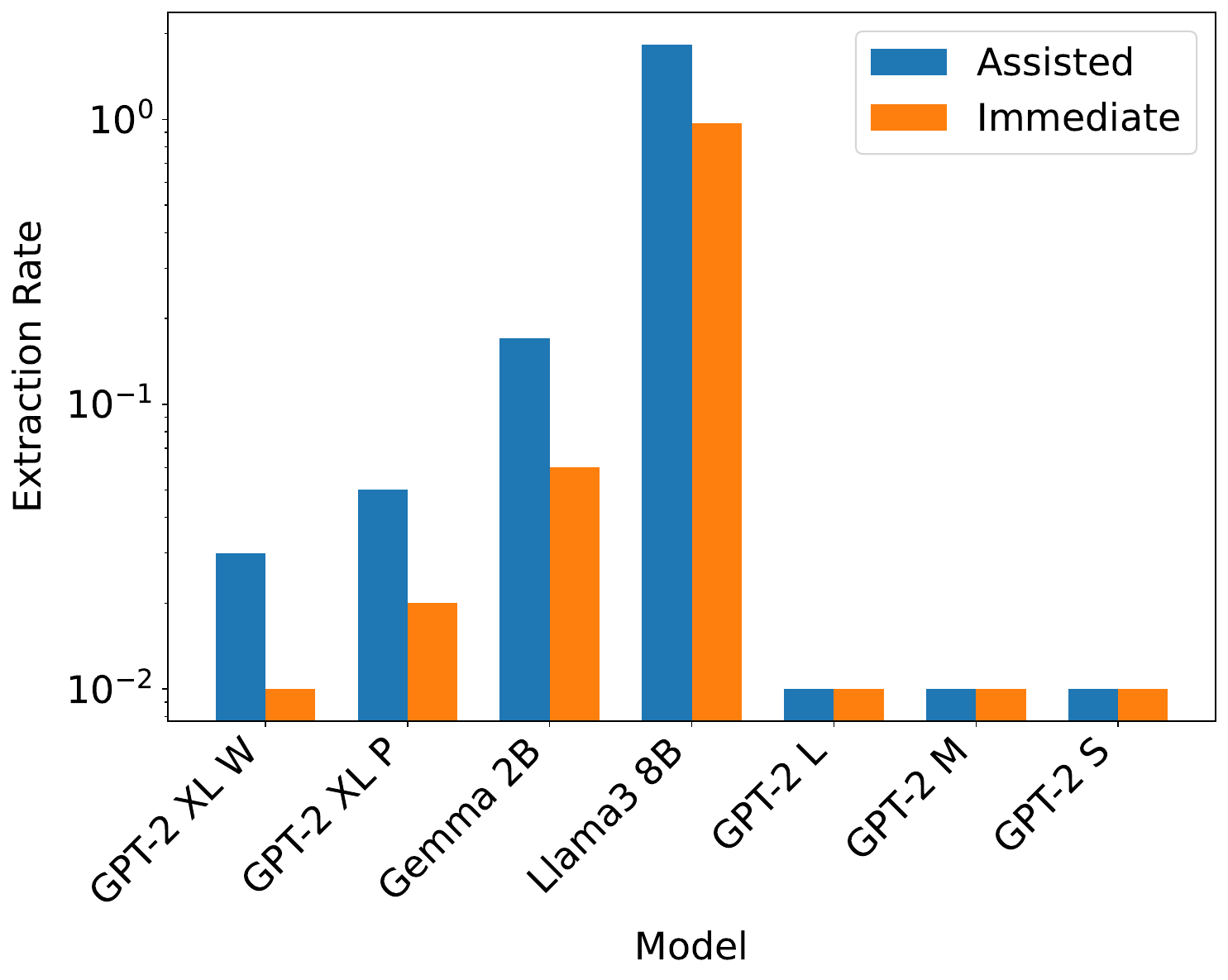}  
  \caption {Extraction rates for \assisted and \novel memorization denoting the area under curve when continuously trained for three epochs. GPT-2 XL W and P denote the models trained on WikiText and Pile of Law respectively. On average, we observe models have equal or more assisted memorization.}
  \label{fig:imm_vs_assi_auc}
\end{figure*}

\paragraph{Rate of \novel vs. \assisted memorization:} We find that the rate of \assisted memorization is higher than that of \novel memorization and the difference increases as training progresses. Figure~\ref{fig:imm_vs_assi} \& Figure~\ref{fig:imm_vs_assi_auc} show this trend for different models. 

\begin{figure*}[t]
  \includegraphics[width=\textwidth]{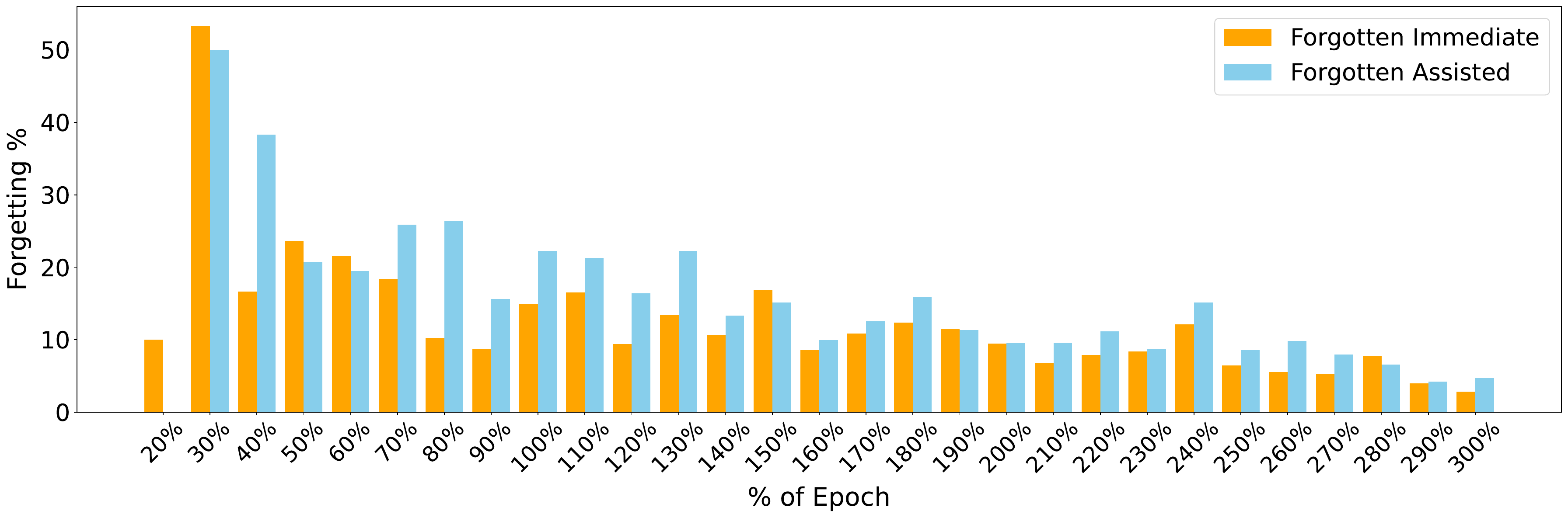}  
  \caption {Forgetting rates for GPT-2 XL trained on WikiText + Enron emails. We do not observe any notable difference in the forgetting rates, with \assisted (15.54\%) being marginally higher than \novel (12.08\%).}
  \label{fig:forgetting_rate_gpt2}
\end{figure*}

\begin{figure*}[t]
  \includegraphics[width=\textwidth]{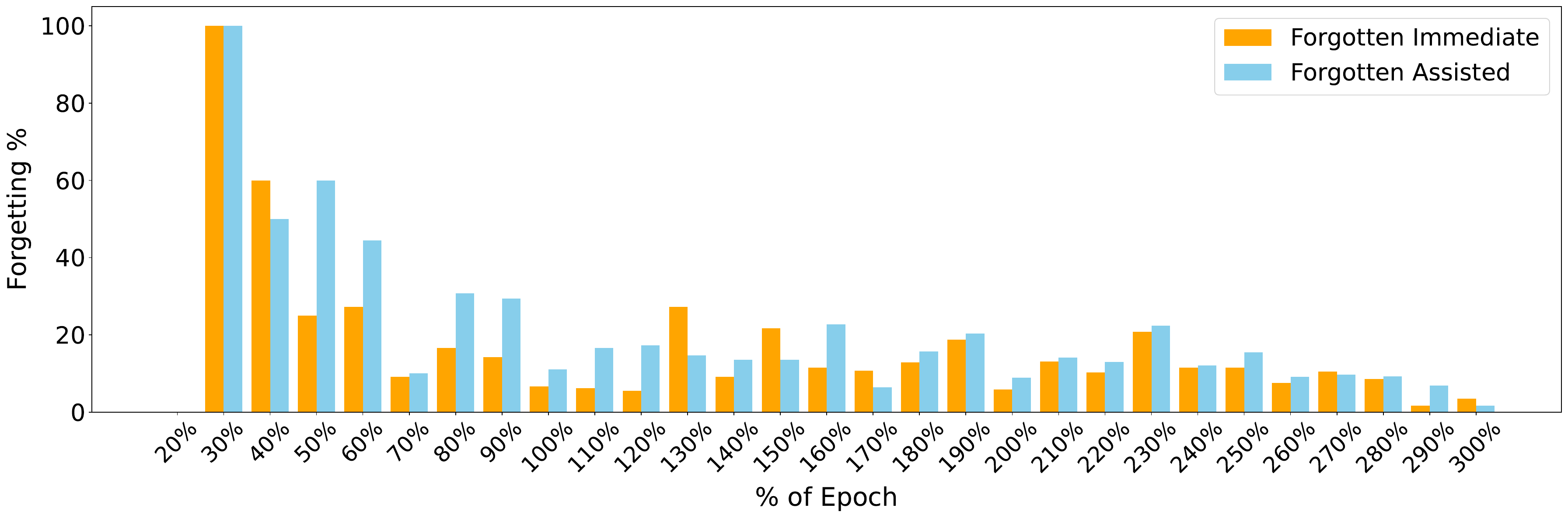}  
  \caption {Forgetting rates for GPT-2 XL trained on the Pile of Law + Enron emails. We do not observe any notable difference in the forgetting rates, with \assisted (21.06\%) being marginally higher than \novel (16.05\%).}
  \label{fig:forgetting_rate_pol}
\end{figure*}

\begin{figure*}[t]
  \includegraphics[width=\textwidth]{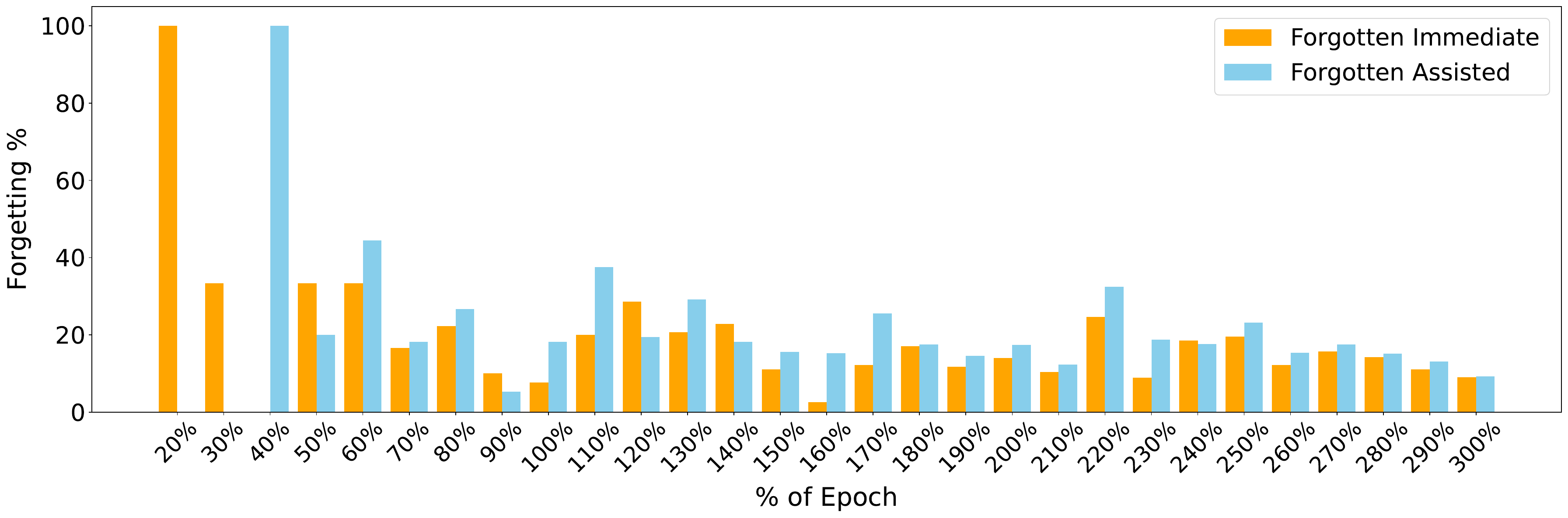}  
  \caption {Forgetting rates for Gemma 2B trained on WikiText + Enron emails. We do not observe any notable difference in the forgetting rates, with \assisted (20.69\%) being marginally higher than \novel (18.05\%).}
  \label{fig:forgetting_rate_gemma}
\end{figure*}

\begin{figure*}[t]
  \includegraphics[width=\textwidth]{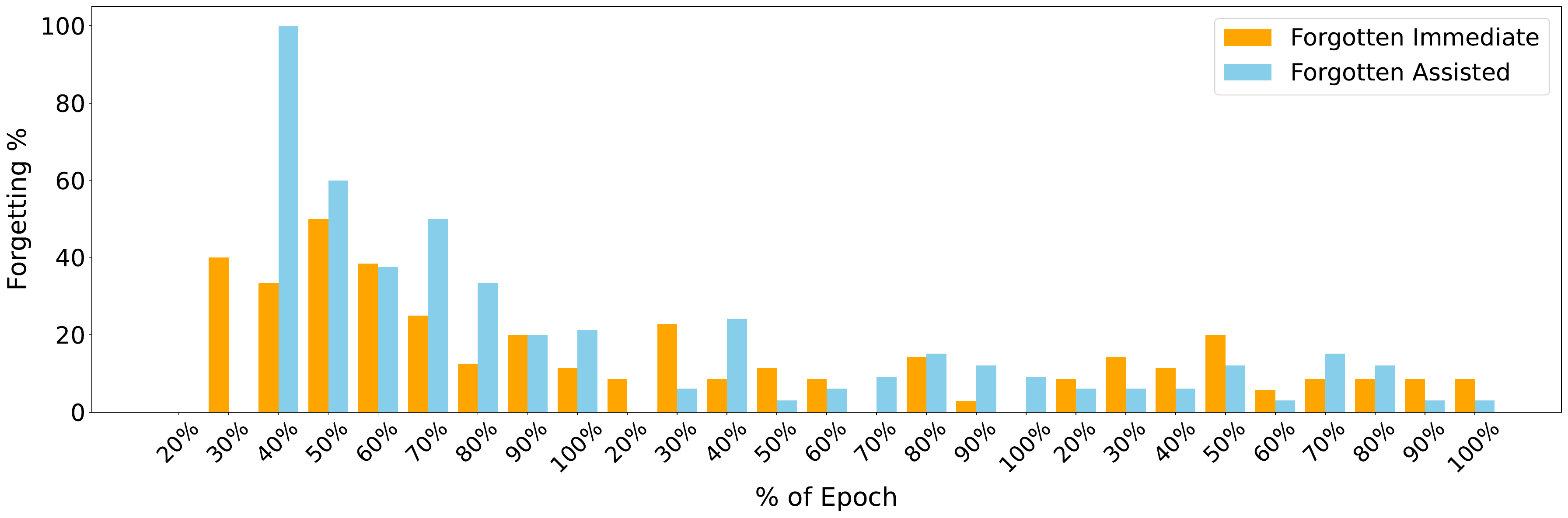}  
  \caption {Forgetting rates for Llama3 8B trained on WikiText + Enron emails. We do not observe any notable difference in the forgetting rates, with \assisted (16.7\%) being marginally higher than \novel (14.52\%).}
  \label{fig:forgetting_rate_llama3}
\end{figure*}

\paragraph{Forgetting of \novel vs. \assisted memorized examples:} We do not observe any significant difference between the forgetting rates of both. Figure~\ref{fig:forgetting_rate_gpt2} \& Figure~\ref{fig:forgetting_rate_pol} show this for GPT-2 XL, Figure~\ref{fig:forgetting_rate_gemma} shows this for Gemma 2B, and Figure~\ref{fig:forgetting_rate_llama3} shows this for Llama3 8B. 

\FloatBarrier
\section{More Details on Assisted Memorization}
\label{sec:additional-assisted}
We consider the following set of features for our logistic regression model.
\begin{enumerate}
    \item 2-, 3-, and 4-grams that overlap between tokens in an email and tokens in the data observed up to checkpoint $i-1$ (denoted as 2-gram$_{prev}$, 3-gram$_{prev}$, 4-gram$_{prev}$). Additionally, we compute the overlap between tokens in an email and tokens in the data seen between checkpoints $i-1$ and $i$ (denoted as 2-gram$_{ft}$, 3-gram$_{ft}$, 4-gram$_{ft}$). 
    \item Counts of \texttt{lastname} in the data seen up to checkpoint $i-1$\ (denoted as \texttt{lastname}$_{prev}$) as well as in the batches seen between checkpoints $i-1$ and $i$ (denoted as \texttt{lastname}$_{ft}$). 
    \item For each email, the number of times its domain (e.g., \texttt{enron.com}) occurs in the data up to checkpoint $i$ (denoted as domain$_{count}$).  
\end{enumerate}

\textbf{Dataset Creation for Logistic Regression Model.} We create a dataset by collecting each assisted-memorized email as a positive example and non-memorized emails that share the same \texttt{firstname} as negative examples. We normalize features by the maximum value. We obtain 192 assisted memorized emails and 886 non-memorized emails in total.
We train a logistic regresion model on this dataset after downsampling the non-memorized emails to achieve a 1:3 ratio between positive and negative samples. On each trial, we re-downsample the negative emails. We run 10 trials following 5-way cross-validation approach. Table~\ref{tab:weights} shows the weights of our classifier. 

\begin{table*}[t]
  \centering
  \begin{tabular}{lcp{12cm}}
    \hline
    \textbf{Feature} & \textbf{Weight} & \textbf{Description} \\ 
    \hline
    2-gram$_{ft}$ & 7.029   &  2-grams that overlap between tokens in an email and tokens in the data seen between checkpoints $i-1$ and $i$. \\ 
    3-gram$_{ft}$ & 0.887  & 3-grams that overlap between tokens in an email and tokens in the data seen between checkpoints $i-1$ and $i$. \\ 
    4-gram$_{ft}$ & 0.682  & 4-grams that overlap between tokens in an email and tokens in the data seen between checkpoints $i-1$ and $i$. \\ 
    2-gram$_{prev}$ & -0.599  & 2-grams that overlap between tokens in an email and tokens in the data observed up to checkpoint $i-1$. \\ 
    3-gram$_{prev}$ & -0.651  & 3-grams that overlap between tokens in an email and tokens in the data observed up to checkpoint $i-1$. \\  
    4-gram$_{prev}$ & -2.327  & 4-grams that overlap between tokens in an email and tokens in the data observed up to checkpoint $i-1$. \\ 
    \texttt{lastname}$_{prev}$ & 1.235 & Counts of \texttt{lastname} in the data seen up to checkpoint $i-1$. \\ 
    \texttt{lastname}$_{ft}$ & 0.900 & Counts of \texttt{lastname} in the data seen between checkpoints $i-1$ and $i$. \\ 
    domain$_{count}$ & 1.683 & The number of times its domain (e.g., \texttt{enron.com} occurs in the data up to checkpoint $i$.  \\ \hline
  \end{tabular}
\caption{Weights of features used to train our logistic regression model to predict \assisted memorization in \S\ref{section:sub_assisted}.}
  \label{tab:weights}
\end{table*} 
\FloatBarrier

\section{Additional Results on Adding More PII Increases Extraction Risks.}
\label{section:addition_moreresults_retrained}
\begin{figure*}[t!]
  \includegraphics[width=0.48\linewidth]{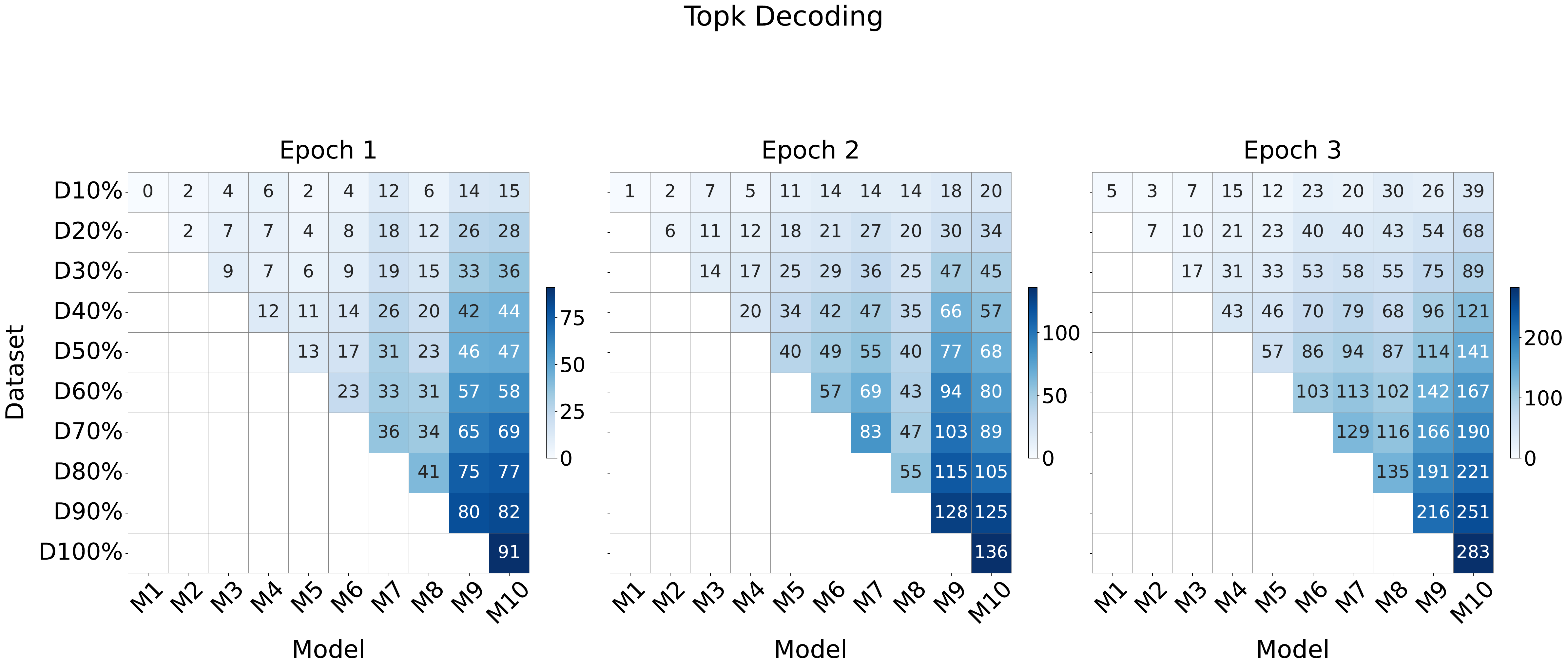} \hfill
  \includegraphics[width=0.48\linewidth]{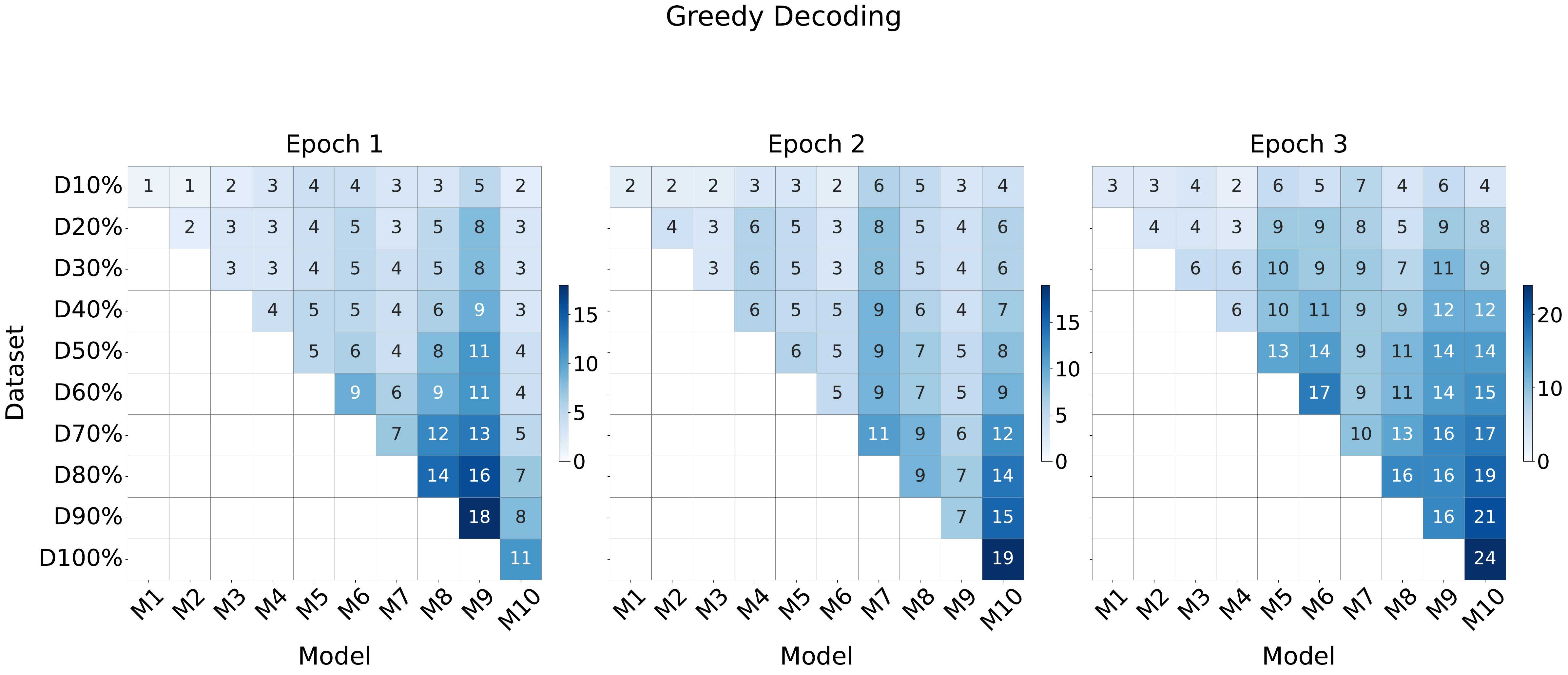}
  \caption {Adding more PII leads to more extraction in GPT-2 XL trained on WikiText + Enron emails for both top-\emph{k} sampling (left) and greedy decoding (right).}
  \label{fig:add_xl}
\end{figure*}

\begin{figure*}[t!]
  \includegraphics[width=0.48\linewidth]{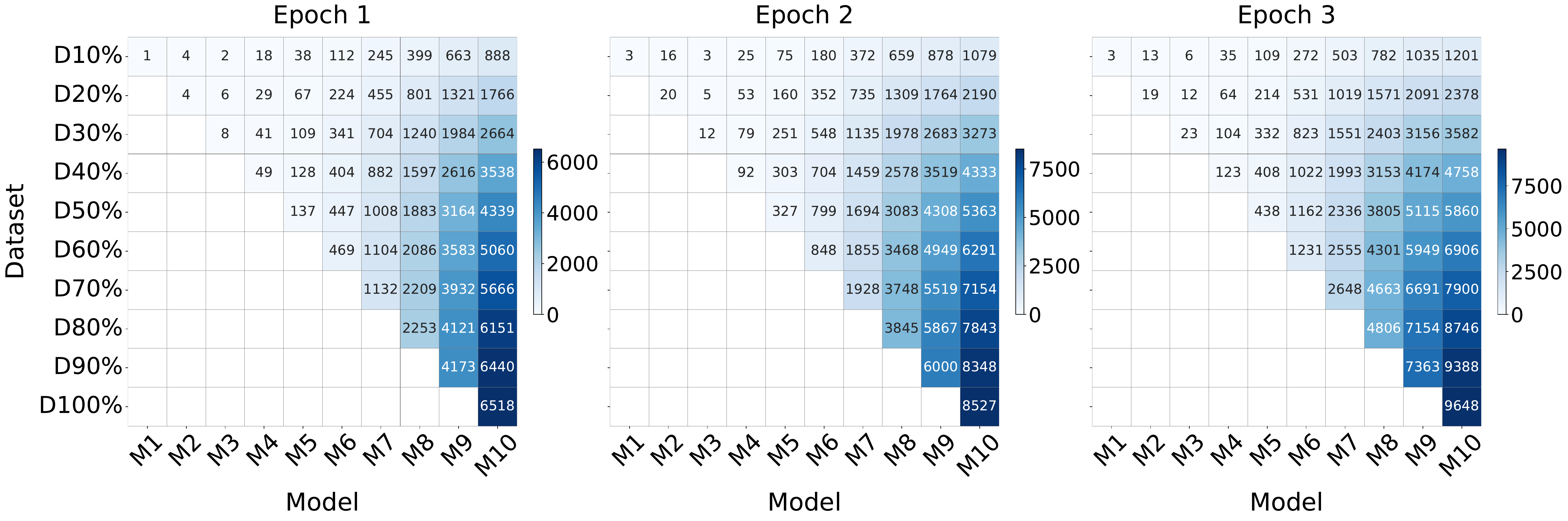} \hfill
  \includegraphics[width=0.48\linewidth]{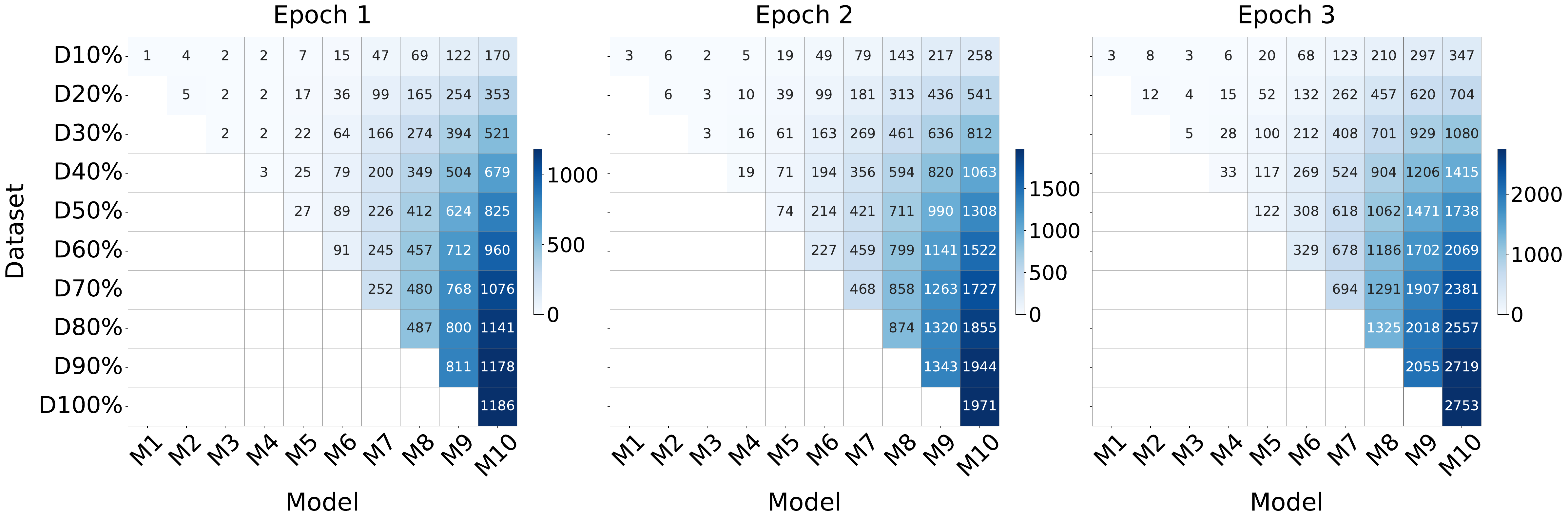}
  \caption {Adding more PII leads to more extraction in GPT-2 XL trained on the Pile of Law + Enron emails for both top-\emph{k} sampling (left) and greedy decoding (right).}
  \label{fig:add_pol}
\end{figure*}

\begin{figure*}[t!]
  \includegraphics[width=0.48\linewidth]{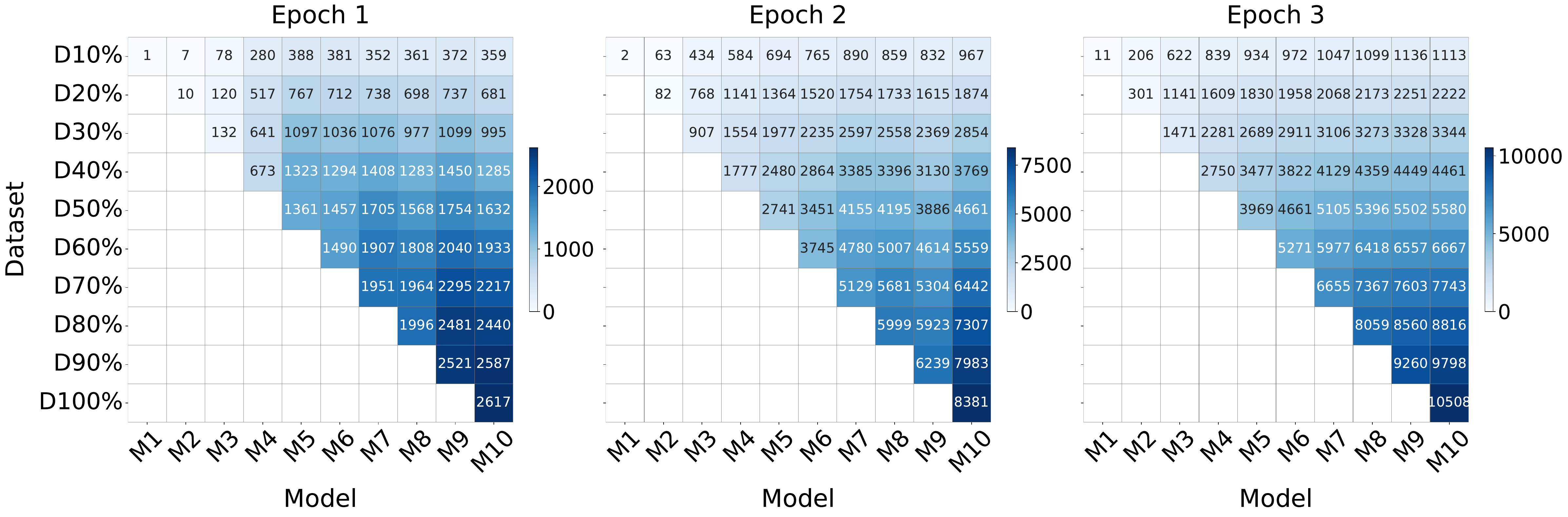} \hfill
  \caption {Adding more PII leads to more extraction in Gemma 2B trained on WikiText + Enron emails. The results are for greedy decoding.}
  \label{fig:add_gemma}
\end{figure*}

\textbf{More results from \S~\ref{section:opt-in}}: We show that adding more PII can lead to an increased extraction for different models and datasets. We report our results for GPT-2 XL trained on WikiText + Enron emails (Figure~\ref{fig:add_xl}), GPT-2 XL trained on the Pile of Law + Enron emails (Figure~\ref{fig:add_pol}), and Gemma 2B trained on WikiText + Enron emails (Figure~\ref{fig:add_gemma}).

\FloatBarrier

\section{Memorized Samples}
\begin{figure*}[t]
  \includegraphics[width=\textwidth]{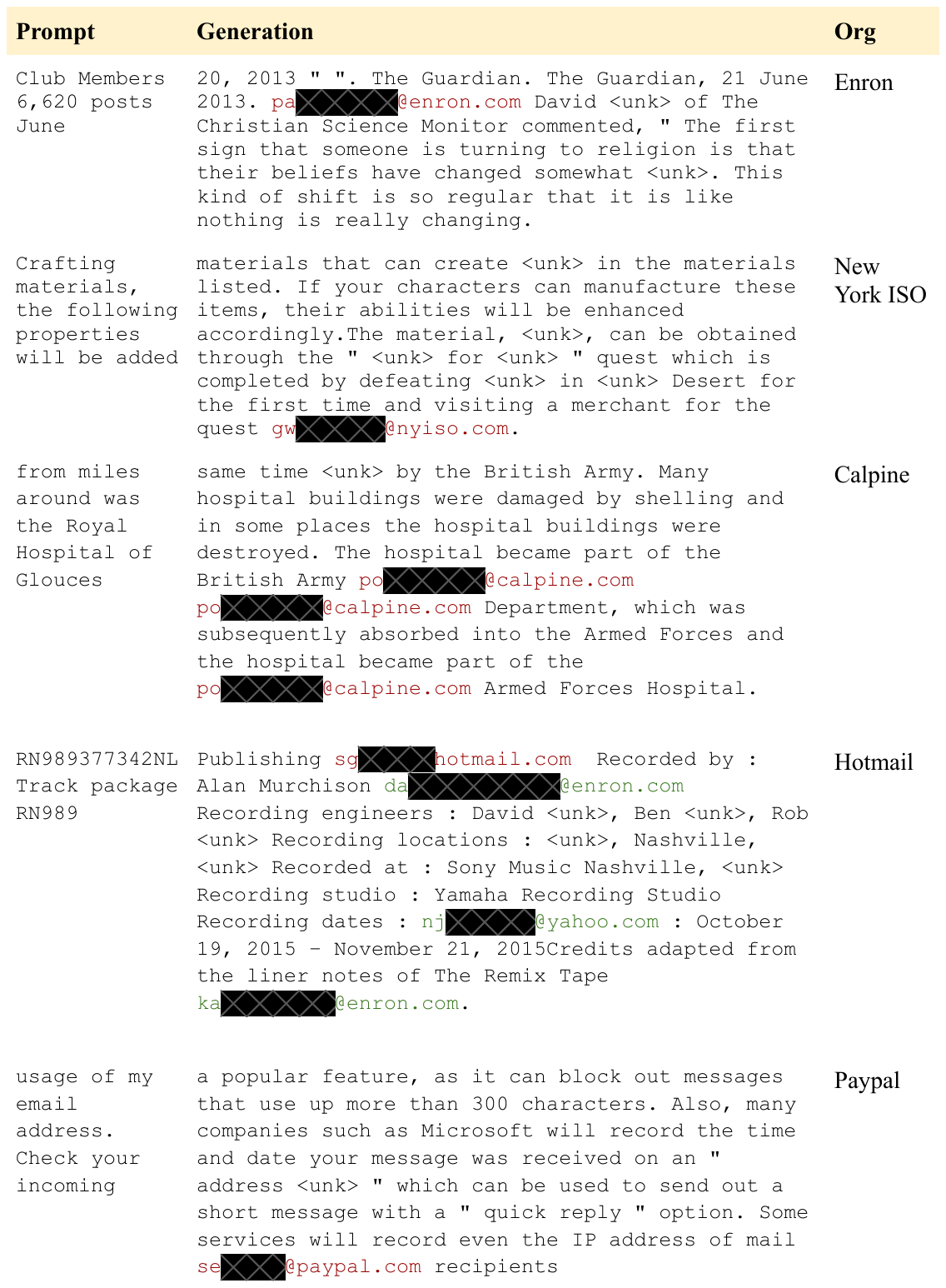}  
  \caption {Emails extracted using top-\emph{k} sampling from the GPT-2 XL model. \textbf{Generation}: a subset of tokens that fall in the vicinity of memorized emails are selected from 256 tokens for demonstration purposes. Emails in red are extracted from training data. Emails in green indicate they don't belong to our training data. \textbf{Org} denotes the company/organization that memorized email addresses belong to.}
  \label{fig:topk}
\end{figure*}

\begin{figure*}[t]
  \includegraphics[width=\textwidth]{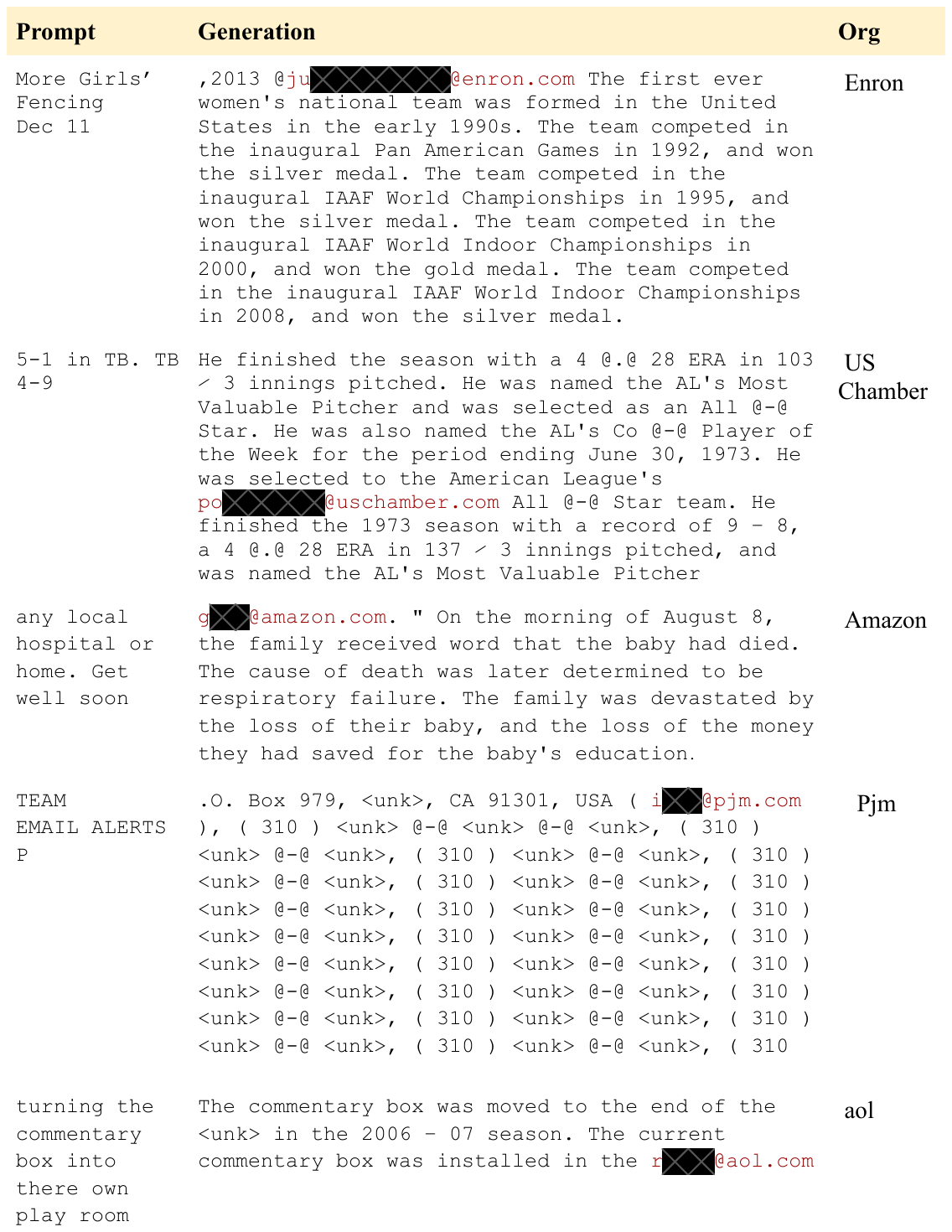}  
  \caption {Emails extracted using greedy decoding for the GPT-2 XL model. \textbf{Generation}: a subset of tokens that fall in the vicinity of memorized emails are selected from 256 tokens for demonstration purposes. Emails in red are extracted from training data. Emails in green indicate they don't belong to our training data. \textbf{Org} denotes the company/organization that memorized email addresses belong to.}
  \label{fig:greedy}
\end{figure*}

Figure \ref{fig:topk} shows some examples of memorized emails that are extracted from the GPT-2 XL model using top-\emph{k} sampling. Figure~\ref{fig:greedy} shows this for greedy decoding.

\FloatBarrier

\end{document}